\documentclass[10pt,journal,compsoc]{IEEEtran}
\usepackage[utf8]{inputenc}

\usepackage{graphicx}
\usepackage{amsmath}
\usepackage{amssymb}
\usepackage{booktabs}

\usepackage{enumitem}

\usepackage{balance}

\usepackage[pagebackref,breaklinks,colorlinks]{hyperref}

\usepackage[capitalize]{cleveref}
\crefname{section}{Sec.}{Secs.}
\Crefname{section}{Section}{Sections}
\Crefname{table}{Table}{Tables}
\crefname{table}{Tab.}{Tabs.}

\usepackage[utf8]{inputenc} 
\usepackage[T1]{fontenc}    
\usepackage{hyperref}       
\usepackage{url}            
\usepackage{booktabs}       
\usepackage{amsfonts}       
\usepackage{nicefrac}       
\usepackage{microtype}      
\usepackage{xcolor}         
\usepackage{multicol}
\usepackage{hyperref}       
\usepackage{url}            
\usepackage{multirow}

\usepackage{booktabs}       
\usepackage{nicefrac}       
\usepackage{wrapfig}

\usepackage[utf8]{inputenc}
\usepackage{amsmath}
\usepackage{mathtools}
\usepackage{amssymb}
\usepackage{color}
\usepackage{caption,subcaption}

\usepackage[pagebackref,breaklinks,colorlinks]{hyperref}

\ifCLASSOPTIONcompsoc
  \usepackage[nocompress]{cite}
\else
  \usepackage{cite}
\fi

\begin{document}

\title{Refining 3D Human Texture Estimation
from a Single Image} 

\author{Said Fahri Altindis,
Adil Meric,
Yusuf Dalva,
U\u{g}ur G\"{u}d\"{u}kbay,
Aysegul Dundar
\IEEEcompsocitemizethanks{
\IEEEcompsocthanksitem S. F. Altindis, A. Meric, Y. Dalva, U. G\"{u}d\"{u}kbay, A. Dundar are with the Department of Computer Science, Bilkent  University,
Ankara, Turkey.
}

}

\IEEEtitleabstractindextext{
\begin{abstract}
Estimating 3D human texture from a single image is essential in graphics and vision. It requires learning a mapping function from input images of humans with diverse poses into the parametric (\textit{uv}) space and reasonably hallucinating invisible parts. To achieve a high-quality 3D human texture estimation, we propose a framework that adaptively samples the input by a deformable convolution where offsets are learned via a deep neural network.  Additionally, we describe a novel cycle consistency loss that improves view generalization. We further propose to train our framework with an uncertainty-based pixel-level image reconstruction loss, which enhances color fidelity. We compare our method against the state-of-the-art approaches and show significant qualitative and quantitative improvements.  Code and additional results: \href{https://github.com/saidaltindis/RefineTex}{https://github.com/saidaltindis/RefineTex}
\end{abstract}

}

\maketitle

\IEEEdisplaynontitleabstractindextext
\IEEEpeerreviewmaketitle

\section{Introduction}

Estimating 3D human texture from a single image is fundamental in many areas, such as virtual reality (VR), augmented reality (AR), gaming, robotics, and clothes try-on. This problem is very challenging given the requirement for predicting the textures of invisible human body parts and the diversity of the pose and appearance of human bodies.

Predicting a three-dimensional (3D) human textured model from a single image receives increasing attention from the research community. However, many of the proposed methods require labor-intensive, expensive data for training, such as 3D scanning~\cite{huang2020arch, lazova2019360, natsume2019siclope, saito2019pifu} or dense human pose estimation~\cite{lazova2019360, neverova2018dense}. In this work, we aim to learn single-image reconstruction without the expensive 3D labels by relying on only multi-view images~\cite{kanazawa2018end,kolotouros2019learning,xu20203d,zhao2020human,xu2021texformer}. Among existing approaches, Zhao et al.~\cite{zhao2020human} propose to use cross-view consistency to enforce the rendered image to match the image from a different view. Wang et al.~\cite{wang2019re}  incorporate the re-identification loss to train the 3D human texture estimation model. Xu and Loy~\cite{xu2021texformer} set an attention-based architecture to allow information processing globally.

\begin{figure}
    \centering
    \includegraphics[width=1.0\linewidth]{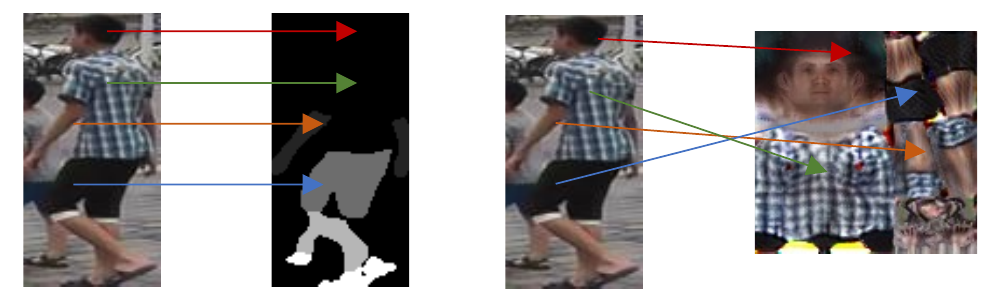}
    \caption{Unlike many tasks with input-output alignment (e.g. part segmentation), texture estimation is not aligned.}
\label{fig:teaser}
\end{figure}

Even though significant progress has been achieved in this domain, previous works still have limitations that hinder the quality of 3D human texture estimation. Firstly, texture estimation from a single image has a set-up that input and output images are not spatially aligned and, therefore, not suitable to solve with Convolutional Neural Networks (CNNs) with local receptive fields. For example, hands can appear anywhere in input images, but they have a fixed corresponding location in the parametric \textit{uv} space (cf.~Fig.~\ref{fig:teaser}). Previous works propose an attention-based architecture to remedy this problem by effectively distributing input features into suitable locations in the parametric \textit{uv} space. 
Our work shows that we can further improve texture estimation via a deformable convolution-based module, which we refer to as the \textit{refinement} module. The learnable offsets of the deformable convolution come from a deep attention-based architecture; therefore, the refinement module can adaptively sample input images to output high-fidelity texture predictions for visible and invisible pixels. Secondly, previous approaches avoid using pixel-level reconstruction loss between rendered and ground-truth images since it performs poorly in generating details. This is because the inaccurate estimation of human body poses and shapes result in misalignments between the rendered and input images.
We propose using a confidence-based pixel-level reconstruction loss to handle the misalignments, significantly improving results. Finally, we enhance our texture estimation with a novel cycle consistency loss.
We apply cycle consistency by estimating texture from a single image, rendering it from a novel view, and encoding the texture again, making the model generalizable to different views. 

We evaluate our method against the current state-of-the-art methods by using metrics such as Structural Similarity Index Measure (SSIM)~\cite{wang2004image}, Learned Perceptual Image Patch Similarity (LPIPS)~\cite{zhang2018unreasonable}, and Cosine Similarity (CosSim)~\cite{sun2018beyond} by comparing the input and rendered images from the same viewpoint, which provides a comparison for the original view. However, when used solely, this evaluation misses the essence of 3D models, which aim to predict appearance from {\em novel} viewpoints. A good model should be able to predict invisible regions with high quality. In our work, we also evaluate the methods from novel views, and we aim to improve the results for the same and other novel viewpoints.

Our contributions are as follows:
\begin{enumerate}

    \item We introduce a deformable convolution-based framework to handle the challenges of mapping unaligned spatially diverse input images into fixed parametric \textit{uv} maps. 
    \item We propose a confidence-based pixel-level reconstruction loss to handle the misalignments in the ground truth. We enable training with pixel-level reconstruction loss and facilitate a closer color appearance to the input image.
    
    \item We design a novel cycle consistency loss that improves the texture estimation quality.
    
\end{enumerate}

We perform extensive evaluations with an array of quantitative metrics that show the effectiveness of our scheme compared to the several state-of-the-art approaches.

\begin{figure*}
    \centering
    \includegraphics[width=0.8\linewidth]{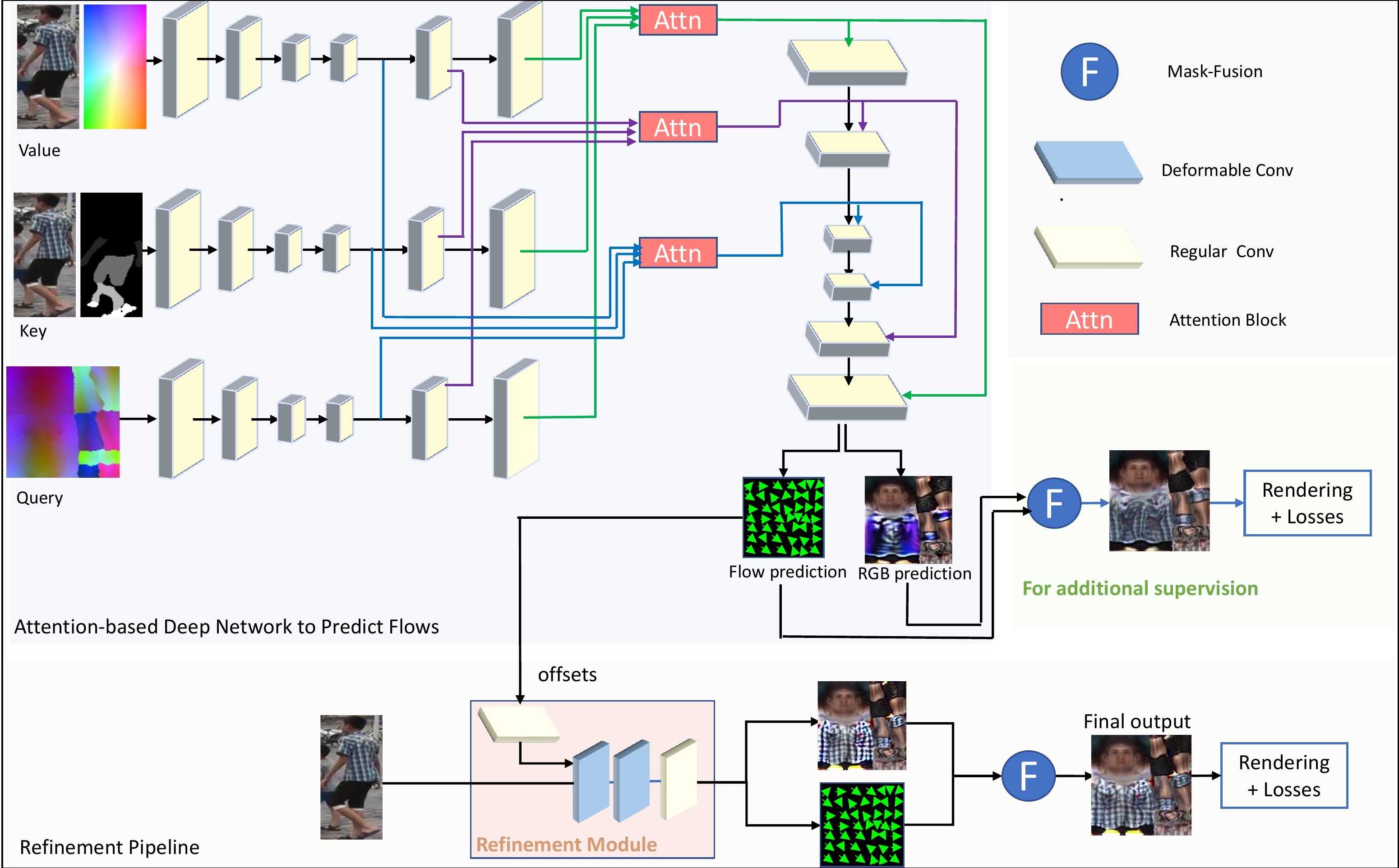}
    \caption{The overall framework. We introduce a deformable convolution-based refinement module where offsets are learned via an attention-based deep network~\cite{xu2021texformer}. This framework can handle the challenges of mapping unaligned spatially diverse input images into fixed parametric \textit{uv} coordinates. We supervise the network with a branch depicted for additional supervision. In this way, offsets receive additional direct supervision as well. Mask prediction for the mask-fusion step is omitted from the figure for brevity.}
\label{fig:overall_frame}
\end{figure*}

\section{Related Work}

3D texture estimation from images can enable various VR/AR applications and attracts much interest from the research community~\cite{kanazawa2018learning,chen2019learning, goel2020shape, li2020self, bhattad2021view,dundar2022fine} especially the 3D human texture estimation \cite{alldieck2018detailed, alldieck2019learning, alldieck2018video, bhatnagar2019multi, huang2020arch, lazova2019360, mir2020learning, saito2019pifu, wang2019re, zhao2020human, zhi2020texmesh, xu2021texformer}. Many methods have been proposed for 3D human reconstruction that take multi-view images and optimize them~\cite{peng2021neural,su2021nerf, jiang2022neuman}. In this work, we are also interested in 3D human reconstruction but inferring them from single-images~\cite{alldieck2019tex2shape, zheng2019deephuman, grigorev2019coordinate, zhao2020human, zhi2020texmesh, xu2021texformer} since it is more applicable to real-world use cases. Even though many works aim to infer texture from single-view images, they may require significantly expensive labor-intensive data during training. For example, most of the methods require 3D scanning~\cite{huang2020arch, lazova2019360, natsume2019siclope, saito2019pifu} such as the recently popularized implicit function-based methods \cite{saito2019pifu, saito2020pifuhd, he2020geo, cao2022jiff} and there are few others that require dense human pose estimation \cite{lazova2019360, neverova2018dense}.
In our work, we aim at learning texture prediction without expensive 3D labels but by relying on image datasets~\cite{wang2019re, zhao2020human, xu2021texformer}.

In this line of research, methods are trained on image collections in a self-supervised manner to reconstruct the input image with a differentiable renderer \cite{kanazawa2018end, kato2018neural, kolotouros2019learning, xu20203d}. 3D body and pose models are predicted with state-of-the-art 3D human mesh reconstruction methods. Additionally, training data is augmented with part segmentation models for superior performance~\cite{zhao2020human, xu2021texformer}. 
While promising progress has been achieved, there are still issues not handled by previous works. One crucial challenge of texture prediction is the lack of strict input and output alignments. The prediction of textures needs to be sampled from a different location for each example. Some methods~\cite{kanazawa2018learning, zhao2020human} learn texture flows to sample pixels from the input image, while some others directly learn the texture maps in RGB values~\cite{wang2019re}, and others use a combination \cite{xu2021texformer}. Learning texture flows can transfer fine details from input to the texture directly, whereas predicting RGB texture can more pleasantly synthesize the invisible regions visually.
The method of Xu and Loy~\cite{xu2021texformer} combines the benefits of these two by learning a fusion strategy. We combine the two approaches in our work with a deformable convolution~\cite{dai2017deformable}. In this way, the prediction of the texture maps becomes easier for the network, which can operate on adaptive offsets.
Deformable convolution is successfully integrated into many computer vision tasks such as object detection, instance segmentation \cite{zhu2019deformable, dai2017deformable}, and texture synthesis \cite{mardani2020neural}.
We show that this task is also suitable for deformable convolution and great improvements can be achieved with deformable based unique design.

Another challenge of 3D texture estimation is to improve the texture predictions for invisible regions. For this problem, previous works utilize multi-view images for cross-view consistency learning~\cite{zhao2020human}. We further improve the results with a novel cycle consistency loss. Last but not least, previous works avoid using reconstruction losses between the input and output due to its degradation in performance~\cite{wang2019re, xu2021texformer}. We propose to use a confidence-based reconstruction loss to improve the results further. 

\section{Method}


\subsection{Architecture}

We work on a set-up where a 3D human texture estimation network is expected to map spatially variant input into the predefined \textit{uv} space coordinates. For example, hands can be anywhere in input images, but they are registered to a fixed coordinate in parametric \textit{uv} space. Since inputs and outputs are not aligned as in most computer vision tasks, the network architecture for this task requires special care. The relevant pixels for synthesizing a coordinate in the parametric \textit{uv} space may be far away in the input space. While we can directly copy visible pixels from the input image to \textit{uv} maps after the locations of corresponding pixels are detected, the network needs to synthesize invisible pixels by conditioning the visible parts. Therefore, we can summarize the goal of the network as two; 1) finding correct offsets for the visible regions so that RGB values from input images can be directly copied, 2) hallucinating RGB values for invisible pixels in the input image by processing what is visible.

We design our framework to have the ability to adapt to the geometric variations in the input image. Our initial goal is to find the offsets of the associated pixels for each pixel in the parametric \textit{uv} map.
Our second goal is to process these associated pixels to output texture predictions with the offsets. The process of finding offsets is similar to flow predictions. However, instead of only sampling the input image with offsets to the output, we want a robust architecture that can adaptively sample input images to process further for better fidelity. 
With this motivation, we employ deformable convolutions. Unlike previous architectures that use deformable convolution, the pathway for predicting offsets includes a deep convolutional neural network, as given in Fig.~\ref{fig:overall_frame}. Previous architectures proposed for texture estimation can be used here as the deep network to predict flow predictions.
We use an attention-based architecture~\cite{xu2021texformer} as shown in Fig. \ref{fig:overall_frame}.
This architecture uses  a color-encoding of the output UV space as query, 2D part-segmentation if input together with the input image as the Key, and input image together with  flow field of the image, i.e., 2D coordinates for each pixel as value. 
Multi-scale features are encoded from these inputs with separate encoder-decoder networks.
The encoded value, key, and query features are input into the attention blocks \cite{b26}. 
Key is used to correlate with the Query elements to obtain the attention map for the input.
Attention maps provide global information to distribute input features to output features.
The outputs of attention blocks go through a \emph{Unet}-like architecture, as shown in Fig. \ref{fig:overall_frame} to output flow and RGB predictions.

We refer to the module that contains deformable convolutions as the refinement module, as shown in Fig.~\ref{fig:overall_frame}.
We calculate the offsets from flow predictions since flow predictions are trained to learn the mapping of the input image to the output.
The flow predictions learn the absolute values of input coordinates. 
We transform them into offsets of pixel coordinates and apply a convolutional layer to them.
With the offsets, the deformable convolution operation can sample the input image from far away pixel coordinates and process them to output texture maps.
The refinement module takes the input image with the size of $128\times64$ and outputs features with the dimension of $128\times128$ based on the offsets in $128\times128$ spatial size.
We modify the deformable convolution implementation to take an arbitrary input size concerning the offsets. The offset size indicates what the output dimension will be. 
This provides an intuitive flow from input to output.

We use the mask-fusion method to combine the advantages of RGB texture prediction and texture flow~\cite{xu2021texformer}. Our network's output consists of the RGB texture map $T_{RGB}$, the texture flow $F$, and a fusion mask $M$. $M$ is not included in Fig. \ref{fig:overall_frame} for brevity. The mask-fusion process is as follows: 

\begin{equation}
\label{eq:fusion}
\small
    T = M \odot f_{sample}(F, I) + (1-M)  \odot T_{\textit{RGB}}, 
\end{equation}

\noindent where $f_{sample}$ refers to the bilinear sampling function that samples textures from the input image, $I$, by the flow predictions, $F$, and $M$ is a binary mask. In this way, visible pixels can be taken by the more accurate $f_{sample}(F, I)$, and invisible pixels can be taken from $T_{\textit{RGB}}$.

We supervise the framework by also rendering the intermediate results from the deep network and backpropagating the losses, as shown in Fig.~\ref{fig:overall_frame}. In this way, offsets receive more direct supervision as well.
We progressively rely more on the supervision the refinement module receives from the final output by turning off the additional supervision.

\begin{figure*}
\centering
\includegraphics[width=0.8\linewidth]{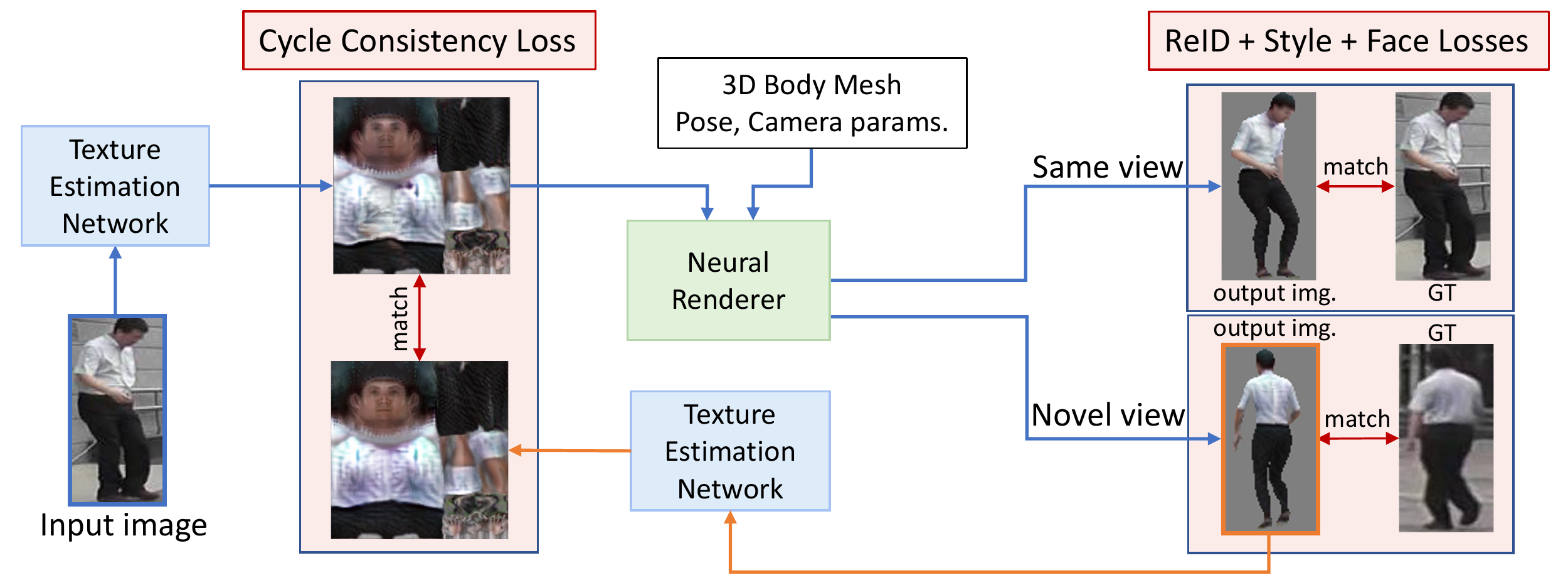}
\caption{We predict a texture from an input image and render it for both the same and novel views. From the image rendered with a novel view, we estimate the texture again. We expect the estimated texture to match the texture predicted from the input image.}
\label{fig:cycle}
\end{figure*}

\subsection{Loss Functions}
\label{sec:baseloss}

The model is trained to minimize various loss functions between the input and rendered images in a self-supervised manner. Given an input image, $I$, our model outputs human texture, $T(I)$. This texture, together with mesh, $M$, and camera, $C$, predictions from state-of-the-art RSC-Net model~\cite{xu20203d} which is built on SMPL model \cite{loper2015smpl} are rendered with a differentiable renderer \cite{kato2018neural}, and the renderer outputs a rendered image, $I^r$. We use the same loss functions between $I$ and $I^r$ as in previous works~\cite{xu2021texformer} and additionally use an uncertainty-based reconstruction loss function and cycle consistency loss, which provide significant improvements. 
This section first reviews the losses we use taken from previous works.
The first loss function is the re-identification loss~\cite{wang2019re}, which minimizes the distance from pedestrian re-identification network ($\Phi$)~\cite{sun2018beyond} at different feature layers ($j$):

\begin{equation}
\label{eq:percep}
\small
    \mathcal{L}_{reid} = ||\Phi_{j}(I) - \Phi_j(I^r) ||^2_2.
\end{equation}

Another previously proposed loss objective for this task is the part-style loss function~\cite{xu2021texformer}. This loss enforces the similarity between each body part of the rendered human and the input image as measured by Gram-matrix~\cite{gatys2016image}. 

\begin{equation}
\label{eq:style}
\small
    \mathcal{L}_{style} = ||G(M_p \odot \Phi_{1}(I)) - G(M^{'}_p \odot \Phi_1(I^r)) ||^2_2, 
\end{equation}

\noindent where $M_p$ and $M^{'}_p$ are the human part segmentation masks from the 2D human parsing model~\cite{huang2018eanet} and 3D mesh, respectively, $p$ indicates the body part, and this loss is calculated for each body part separately and summed together. $G$ stands for Gram-matrix, features are again encoded by the same pedestrian re-identification network ($\Phi$), and only the first layer is used. 


The last loss function we use is the face structure loss~\cite{xu2021texformer}, which generates accurate and reliable faces. Given that all human faces follow the same structure, face structure loss ensures the similarity between estimated textures and synthetically-generated textures.
Each region of the parametric \textit{uv} map is predetermined; hence the location of the face in the texture is constant. This loss only checks similarities between faces by applying a fixed mask:

\begin{equation}
\label{eq:face}
\small
    \mathcal{L}_{face} = -\dfrac{1}{N} \sum_{i=1}^N s(M_{face} \odot T(I), M_{face} \odot F_{syn}^i),
\end{equation}
\noindent where $\{F_{syn}^{i}\}_{i=1}^N$ is a set of synthetically generated human textures obtained from the synthetic human dataset~\cite{DBLP:journals/corr/Varol0MMBLS17}, $M_{face}$ is a predefined binary mask that indicates the face region on the texture map, and $s$ is a structure-similarity function \cite{wang2004image} that calculates face similarities. The loss function in Eq.~\ref{eq:face} optimizes the network to output face texture predictions to have similar structures as $F_{syn}$. This results in the generation of plausible face textures while retaining the colors of the input human.
The losses we use from previous works are our base losses, and Eq.~\ref{eq:base} gives the overall base loss.

\begin{equation}
\label{eq:base}
\small
    \mathcal{L}_{base} = \lambda_1 \times \mathcal{L}_{reid}  + \lambda_2 \times \mathcal{L}_{style} + \lambda_3 \times \mathcal{L}_{face},
\end{equation}
\noindent where $\lambda_1=5000, \lambda_2=0.4$, and $\lambda_3=0.01$. We take these parameters from previous work and do not tune them.

\subsubsection{Uncertainty-based Reconstruction Loss}

Previous approaches avoid using pixel-level reconstruction loss between rendered and input images since it performs poorly in generating details. The inaccurate estimation of human body poses and shapes may result in misalignments between the rendered and input images. Due to this problem, 3D human texture estimation models are trained only with re-identification losses that compare features at a high level and style losses that do not use spatial correspondence. On the other hand, it is shown that pixel-level reconstruction losses improve results when the generated and output images are aligned~\cite{isola2017image,wang2018high,park2019semantic,dundar2020panoptic}.
To take advantage of pixel-level reconstruction loss and be robust to misalignments, we propose to estimate a confidence map, $\sigma$, to adjust the reconstruction loss objective.
The ground-truth output image, $I$, and the rendered image, $I^r$, are compared via the loss given in Eq. \ref{eq:uncertainty_recon} as was also defined in \cite{wu2020unsupervised}.

\begin{equation}
\label{eq:uncertainty_recon}
\small
    \mathcal{L}_{url} = -\sum_{x,y} \ln \left( {\frac{1}{\sqrt{2 \sigma_{x,y}^2}}  \exp-\frac{\sqrt{2} |I_{x,y} - I_{x,y}^r |}{\sigma_{x,y}}} \right).
\end{equation}

In this loss objective, the role of the confidence map, $\sigma$, is to estimate the aleatoric uncertainty of the model~\cite{kendall2017uncertainties}. $(x,y)$ are the spatial pixel coordinates of $I$. The objective is the negative log-likelihood of a factorized Laplacian distribution with the mean predicted by the model and $\sigma$ predicted by the confidence model. This way, the model calibrates itself and minimizes the reconstruction loss by optimizing the confidence map~\cite{kendall2017uncertainties,wu2020unsupervised}.
The confidence model has the \textit{Unet} architecture~\cite{ronneberger2015u}, which takes the input image, $I$, and outputs the $\sigma$. This model is only used during training and is not needed during inference.

\subsubsection{Cycle Consistency Loss} We also propose a cycle loss to enforce consistency between the textures estimated by the model from input images and the model's renderings for different views. In the cycle process, the model generates the initial texture map, $\textit{uv}_{\textit{map}}$, from the given input image as $I$ by the texture estimation network, $T$. We render a new image using the estimated textures for a different view obtained from another image of the same person with a renderer, $R$, Body Mesh Parameters, $m$, and camera parameters, $c$. From the rendered image, the network estimates the texture again. We expect these two estimated textures to be as close as possible; hence, we calculate a pixel-wise L2 loss between two estimated textures, as given in Eq. \ref{eq:cycle}.

\begin{align}
\label{eq:cycle}
\small
    \mathcal{L}_{cyc} & = ||T(I)  - T(R(T(I), m, c))||_2. 
\end{align}

Fig.~\ref{fig:cycle} explains the detailed structure of the cycle consistency. In previous methods, the texture map is estimated from the input image, and it is rendered for the same and novel views to enforce multi-view consistency and minimize base losses, as described in the previous subsection. We also estimate the texture from the novel view rendering to obtain additional guidance for the texture estimation network. This additional loss also helps us to achieve our second objective, where we can compare images at the pixel level.


\subsubsection{Total Loss} The baseline loss $L_{base}$ defined at Eq. \ref{eq:base} is calculated twice for two different rendered images. The first baseline loss, $L_{base-sv}$, is calculated between the input image and the image rendered with the input image's generated texture and camera parameters. Moreover, to improve texture estimation of unseen parts, the second baseline loss, $L_{base-nv}$, is calculated between the ground-truth image from a different view and the image rendered with the generated texture of the input image and camera parameters of the novel-view ground-truth image. $L_{base-nv}$ provides the multi-view consistency. Additionally, we have two new losses. The overall loss function to train our framework is as follows: 

\begin{equation}
\label{eq:total}
\small
    \mathcal{L}_{total} = \mathcal{L}_{base-sv} + \mathcal{L}_{base-nv} +
    \lambda_4 * \mathcal{L}_{cyc} + \lambda_5 * \mathcal{L}_{url},
\end{equation}

\noindent where $\lambda_4=0.1,$ and $\lambda_5=10^{-3}$.

\begin{table*}[t]
\caption{Comparisons of models trained with multi-view consistency and without multi-view consistency (single-view images). Results in bold indicate the best in each column.}
\label{tab:sview}
\centering
\renewcommand\tabcolsep{12pt}
\begin{tabular}{|l|cc|cc|cc|cc|}
\hline
& \multicolumn{2}{c}{SSIM \ \ \ $\Uparrow$} & \multicolumn{2}{|c}{LPIPS \ \ \ $\Downarrow$}  & \multicolumn{2}{|c}{CosSim \ \ \ $\Uparrow$}  & \multicolumn{2}{|c|}{CosSim-R \ \ \ $\Uparrow$} \\
\hline
& SV & NV & SV & NV & SV & NV & SV & NV \\
\hline
Trained with Multi-view &  0.7422 & \textbf{0.6535} & 0.1154  & \textbf{0.2040} & 0.5747 & \textbf{0.4943} & 0.5422 & \textbf{0.4736}\\
Trained with Single-view & \textbf{0.7706} & 0.6494 & \textbf{0.0963} & 0.2150 & \textbf{0.5823} & 0.4809  & \textbf{0.5496} & 0.4585 \\
\hline
\end{tabular}
\end{table*}

\begin{figure*}
    \centering
    \includegraphics[width=0.8\linewidth]{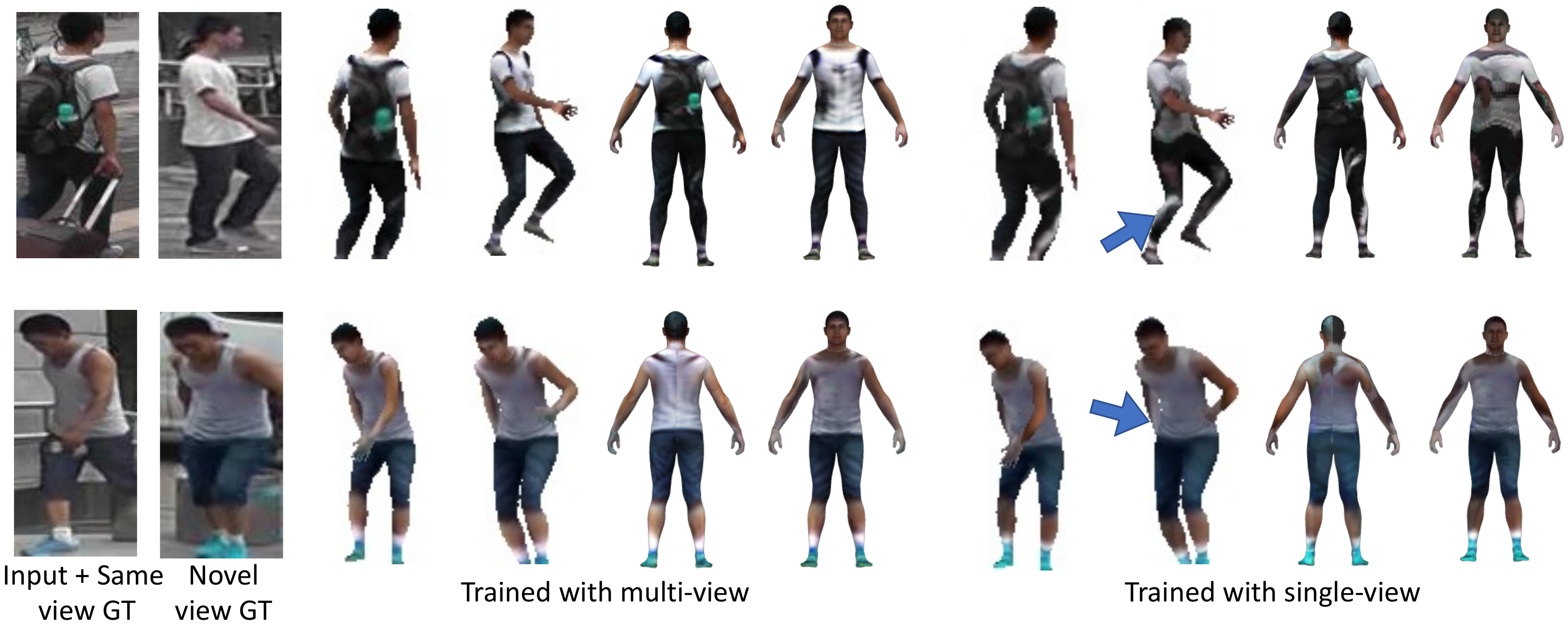}
    \caption{Input/same view and novel view ground-truth images are provided in the first two columns. Other columns show the baseline-trained results with multi-view consistency and trained without multi-view consistency (with single-view images). Blue arrows show that models trained with single-view images better reconstruct the same view but not the novel one. Therefore, instead of only evaluating the models based on the reconstruction of the same view, which misses the point of 3D models, we additionally evaluate the models from a novel view.}
\label{fig:sv_eval}
\end{figure*}

\section{Experiments}
\label{sec:exp}

\textbf{Dataset, Architecture, and Training Details.} We use the Market-1501 dataset \cite{zheng2015scalable} in our experiments. Among the human images of 1501 person identities, we use the same training-testing split as in other works~\cite{wang2019re,xu2021texformer}.
We additionally run experiments on the DeepFashion dataset~\cite{liu2016deepfashion}. 
The DeepFashion dataset includes in-shop clothes images. These images show large pose and view variations. We follow the same train/test split from previous work \cite{zhao2020human}.
Previous work \cite{zhao2020human} removes the images that only contain a small human body part and ends up with 20,185 training and 6,639  testing images.
We use their split.

The overall architecture includes deformable and traditional convolution layers.
Starting with the refinement module, 
a deformable convolution receives initial offsets from a deep attention-based network.
This attention-based network has an encoder-decoder architecture where at each scale, there is an attention block as introduced in \cite{xu2021texformer}.
The network has $6$ convolution layers, each with a filter size of $3\times3$ and $128$ channel size.
There are downsamplings after the second and third convolution layers and upsamplings after the fourth and fifth convolution layers.
The first three convolution layers correspond to the encoder, and the other three belong to the decoder.
There are skip connections between the encoder and the decoder at each scale which employs an attention block, and the output of the attention blocks is summed up with the decoded features.

The initial offsets coming from the deep network go through a convolution layer to output 18 channels ($2\times3\times3$) to be used as the offsets to the deformable convolution layer with filters $3\times3$ width and height and $128$ output channels. 18 channel corresponds to the $x, y$ offsets for each pixel in a kernel ($3\times3$).
The offsets are a dimension of $b\times18\times128\times128$, and input has a dimension of $b\times4\times128\times64$ where channel size of $4$ refers to the RGB image with part segmentation concatenated and $b$ is batch size.
We modify the deformable convolution implementation to take an arbitrary input size with respect to the offsets.
Offset size indicates what the output dimension will be.
Therefore, from the input image with a spatial dimension of $128\times64$, we output feature maps with a spatial dimension of $128\times128$.
The output here is processed with additional deformable and convolutional layers.
For each convolution layer, we use kernel size 3, stride 1, and padding 1. The channels change as \{4, 128, 128, 128, 6\}.
We additionally have a skip connection from the deep attention network's predictions to the refinement module. We concatenate the predictions from the deep network with the deformable convolution's output.
The final channel size of 6 corresponds to the RGB predictions (3 channels), UV flow predictions (2 channels), and fusion mask prediction (1 channel).
The RGB and UV flow predictions go through Tanh activation, and mask predictions go through sigmoid activation layers.
Each convolution layer in the overall architecture is followed by batch normalization and ReLU layers.

The confidence model has $6$ convolution layers, each with a filter size of $3\times3$ and $128$ channel size.
Again, each convolution layer is followed by batch normalization and ReLU layers.
There are downsamplings after the second and third convolution layers and upsamplings after the fourth and fifth convolution layers.
The first three convolution layers correspond to the encoder, and the other three belong to the decoder.
There are skip connections between the encoder and decoder at each scale. Encoded features and decoded features are summed up via skip connections.
The last convolutional layer reduces the channel size to $1$, followed by the softplus activation function.
This network is only used during training.

We use the Adam optimizer and train our framework with a batch size of 16 and a learning rate of $1\times10^{-3}$ and betas=$(0.9, 0.999)$ for 200 epochs.
We do not use a learning rate scheduling and keep the learning rate the same for 200 epochs.

\textbf{Evaluation Metrics.} Following the previous works, we use various metrics for evaluation. SSIM~\cite{wang2004image} and LPIPS~\cite{zhang2018unreasonable} are used to find pixel- and feature-level similarities between the output and ground-truth images.
Additionally, cosine similarities of person ReID features~\cite{sun2018beyond} are computed to evaluate a high-level semantic similarity, e.g., how likely the rendered human is the same person from the ground-truth image. Following previous work~\cite{xu20213d}, we calculate this metric with two different networks, PCB~\cite{sun2018beyond} and TorchReid~\cite{zhou2019torchreid}, resulting in CosSim and CosSim-R metrics, respectively.

These metrics are previously reported only between the input and rendered images from the same view. However, only the visible texture estimates are evaluated since that evaluation protocol only measures the estimated texture from the same view. We also measure the metrics from a novel view for which we have the ground truth to overcome this limitation. Given an input image from one view, we render the estimated 3D human texture based on a camera view and pose estimate from another. We evaluate the results for all the other available views for each person. We refer to the same view evaluation as \textit{SV} and the novel view as \textit{NV} in the tables.

\begin{table*}[t]
\caption{Evaluation results of our method and state-of-the-art competing methods. Results in bold indicate the best in each column, and underscored results indicate the second.}
\label{tab:final}
\centering
\renewcommand\tabcolsep{8pt}
\begin{tabular}{|l|l||cc|cc|cc|cc|c|}
\hline
& & \multicolumn{2}{c}{SSIM \ \ \ $\Uparrow$} & \multicolumn{2}{|c}{LPIPS \ \ \  $\Downarrow$}  & \multicolumn{2}{|c}{CosSim \ \ \ $\Uparrow$}  & \multicolumn{2}{|c|}{CosSim-R \ \ \ $\Uparrow$} & Params. (M)  \\
\hline
Dataset & Method & SV & NV & SV & NV & SV & NV & SV & NV & \\
\hline
\multirow{ 5}{*}{Market-1501}  & HPBTT  \cite{zhao2020human} &  0.7380 & 0.6496 & \underline{0.1148} & 0.2156 & 0.5336 & 0.4697 & 0.5077 & 0.4508 & 42.3\\
& RSTG \cite{wang2019re} &  0.6735 & 0.6283 & 0.1778 & 0.2421 & 0.5282 & 0.4717 & 0.4924 & 0.4454 & 13.4 \\
& TexGlo \cite{xu20213d} & 0.6658 & - & 0.1776 & -& 0.5408 & - & 0.5048 & -  & 16.1\\
& Texformer \cite{xu2021texformer} & \underline{0.7422} & \underline{0.6535} & 0.1154  & \underline{0.2040} & \underline{0.5747} & \underline{0.4943} & \underline{0.5422} & \underline{0.4736} & 7.6\\ 
& Ours  & \textbf{0.7611} & \textbf{0.6544} & \textbf{0.1003} & \textbf{0.2040} & \textbf{0.5858} & \textbf{0.4963} & \textbf{0.5538} & \textbf{0.4758} & 8.2 \\
\hline
\multirow{ 3}{*}{DeepFashion}  & HPBTT  \cite{zhao2020human} & 0.6506 & \textbf{0.6391} & 0.3527 & 0.3669 & 0.6193 & \underline{0.5808} & 0.6122 & \underline{0.5717} & 42.3 \\
& Texformer \cite{xu2021texformer} & \underline{0.7354} & 0.5479 & \underline{0.1173} & \underline{0.3160} & \underline{0.7595} & 0.5536 & \underline{0.7429} & 0.5451 & 7.6 \\ 
& Ours & \textbf{0.8358} & \underline{0.5995} & \textbf{0.0871} & \textbf{0.2689} & \textbf{0.7728} & \textbf{0.6027} & \textbf{0.7582} & \textbf{0.5897} & 8.2 \\
\hline
\end{tabular}
\end{table*}

We experiment with the reliability of novel view results by comparing baseline methods trained with and without multi-view consistency. Multi-view consistency is broadly used on datasets that contain multiple images taken from different views of the same object~\cite{zhao2020human,xu2021texformer}. One view is used as input in these settings, and loss functions are calculated on the same input view and a novel view. The multi-view consistency improves texture estimation quality for the invisible regions because networks receive gradients from the novel image target for invisible parts in the input image. 

Table~\ref{tab:sview} compares our baseline models, trained with multi-view and single-view data, which refers to using multi-view consistency and not, respectively. The traditional evaluation results on the same view show that the single-view baseline model achieves significantly better scores. This is because it is tuned to estimate the visible areas perfectly, producing better results on the same but worse on a novel view, as seen in Fig.~\ref{fig:sv_eval}.
The quality of renderings in novel views is usually more valuable and the primary purpose of 3D models. Table~\ref{tab:sview} shows the evaluation of the quality of novel view renderings. This setup indicates the limitation of previously used evaluation protocol by 3D human texture estimation models. In this work, we are interested in improving the quality of both the same and novel views, and we keep track of both of these metrics. 

\subsection{Comparison with the State-of-the-art}

\newcommand{\interpfigm}[1]{\includegraphics[height=2.4cm]{#1}}
\newcommand{\interpfigt}[1]{\includegraphics[height=3.1cm]{#1}}
\begin{figure*}[]
  \centering
  \setlength\tabcolsep{1.8pt}
  \renewcommand{\arraystretch}{0.48}
  \begin{tabular}{ccccc}
  \interpfigm{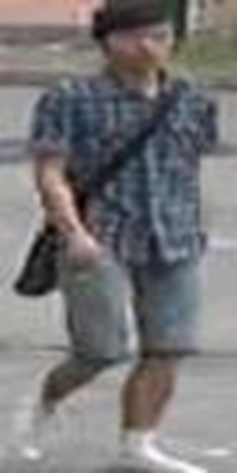}&
    \interpfigt{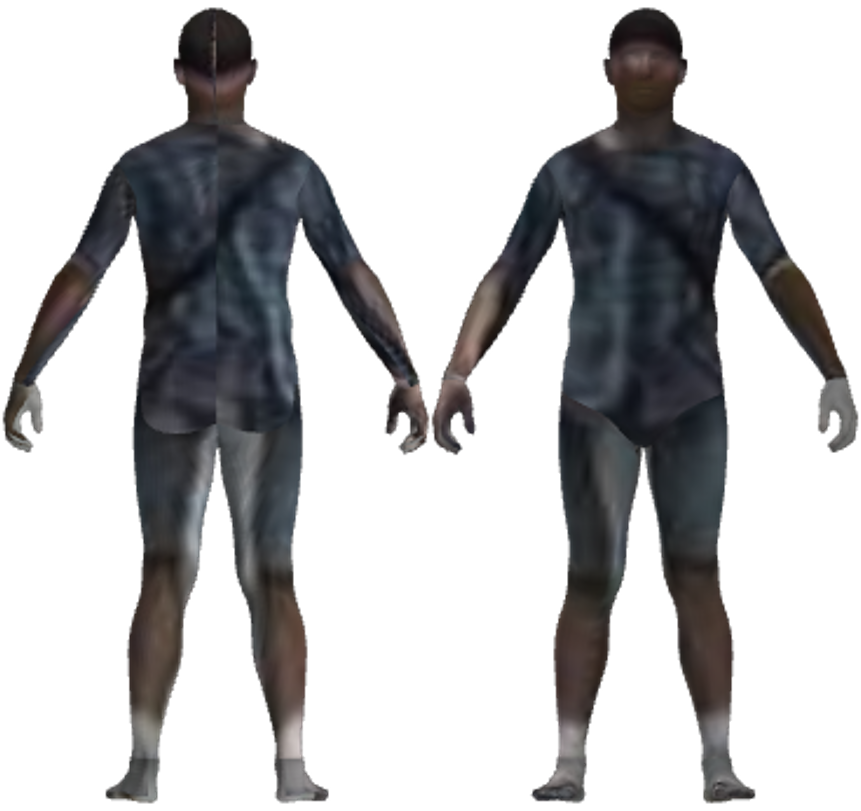}&
    \interpfigt{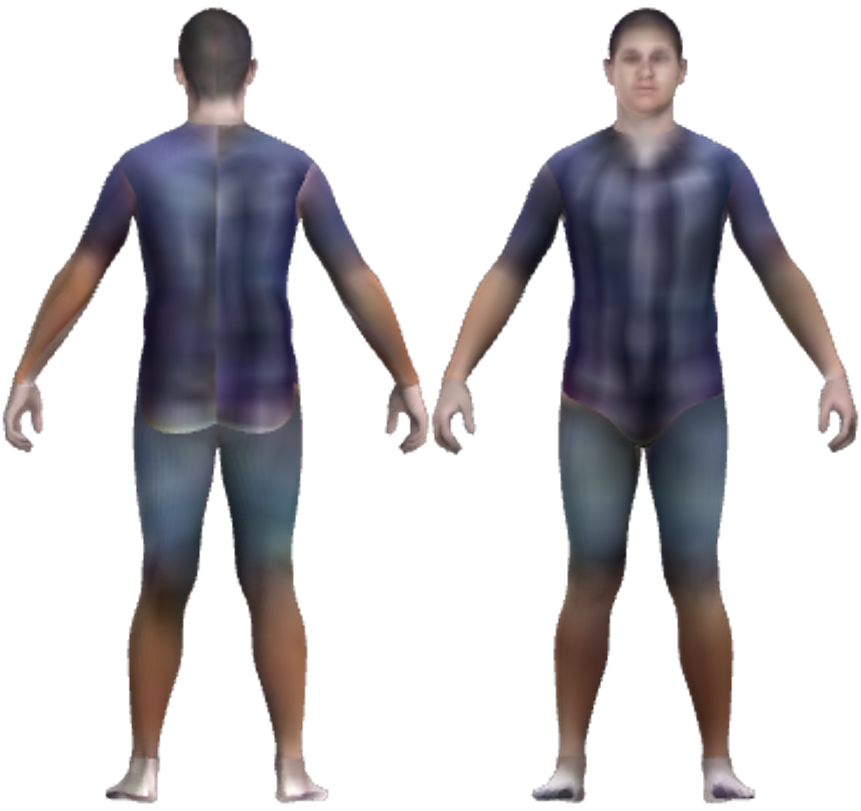}&
    \interpfigt{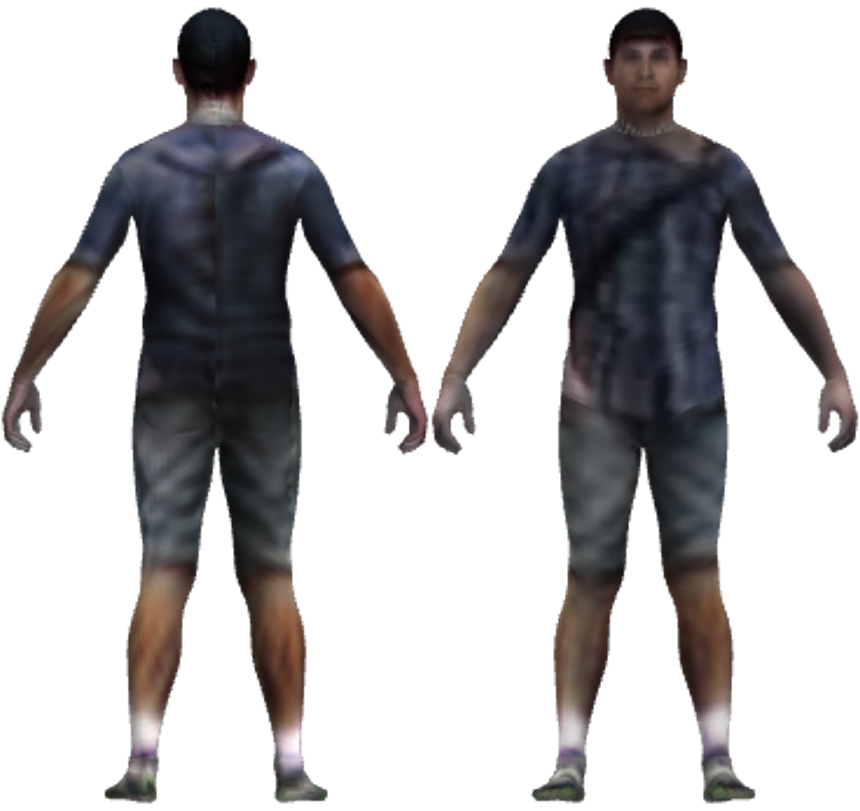}&
    \interpfigt{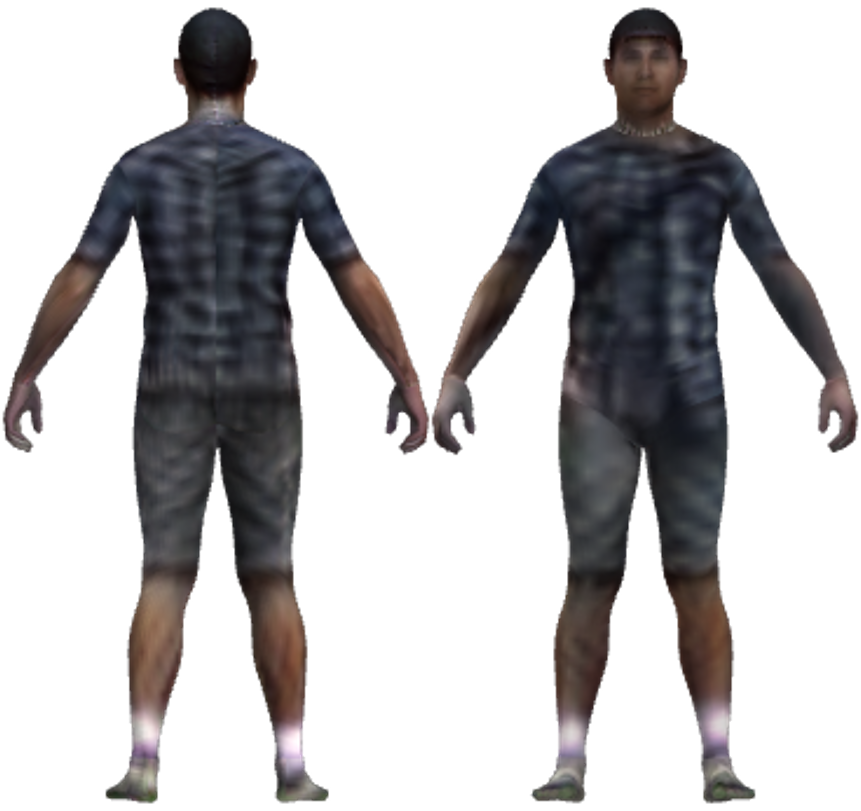}\\

    \interpfigm{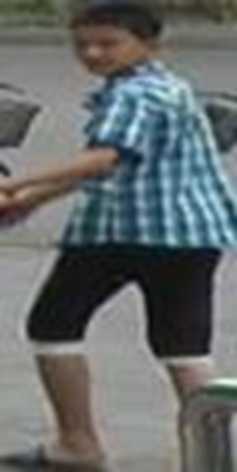}&
    \interpfigt{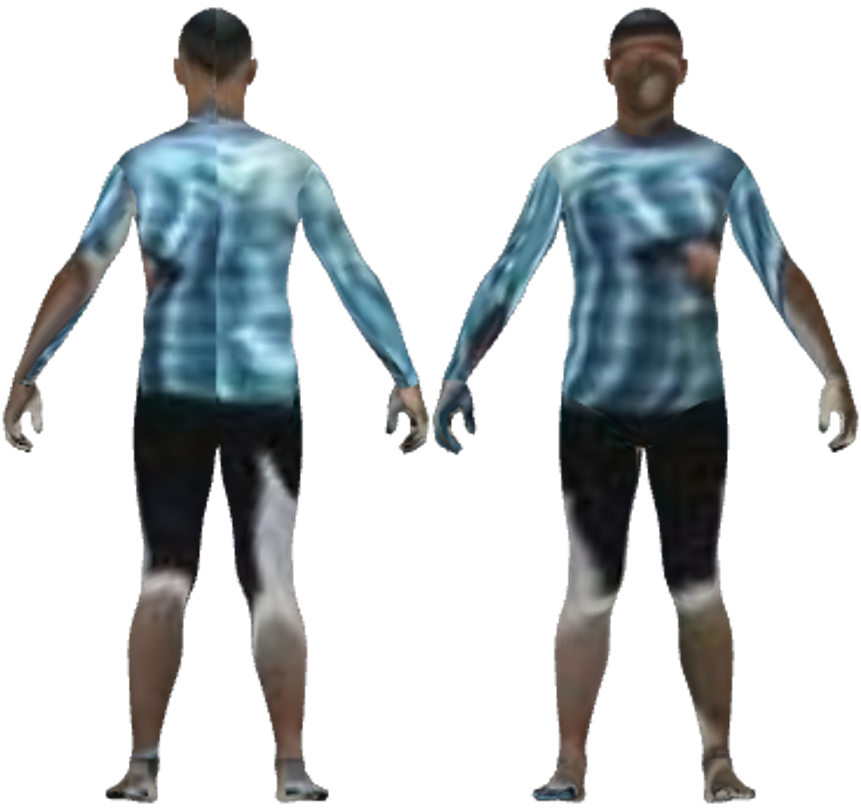}&
     \interpfigt{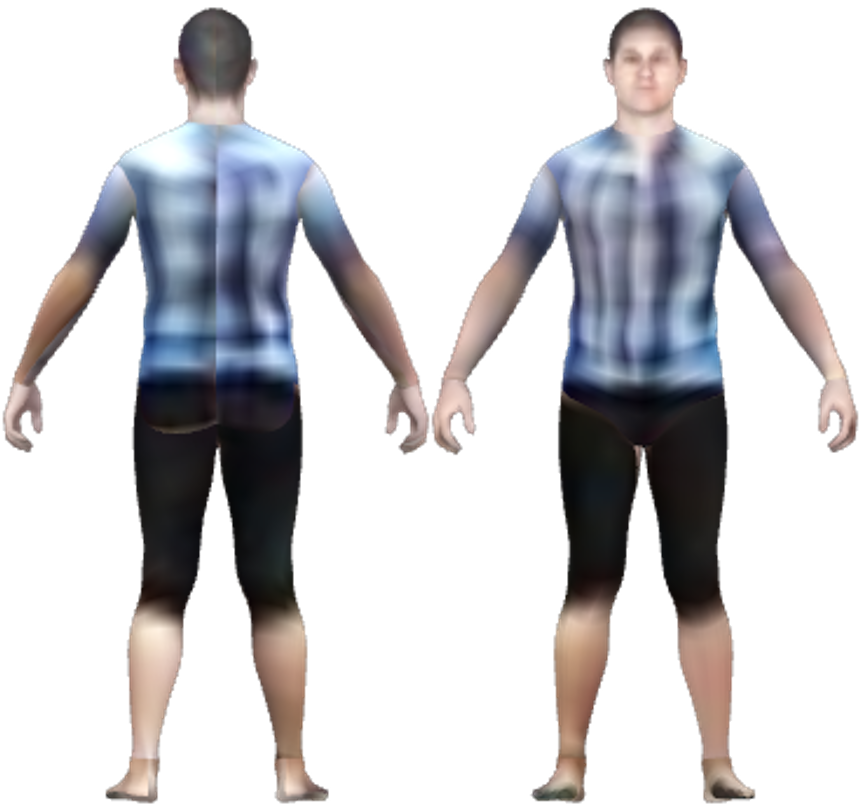}&
     \interpfigt{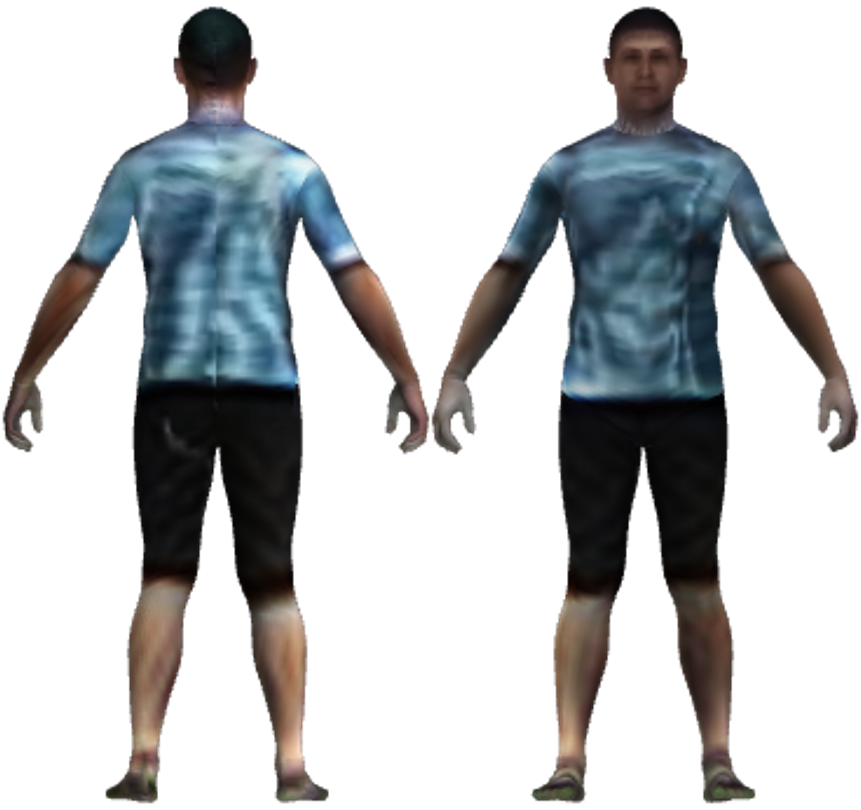}&
     \interpfigt{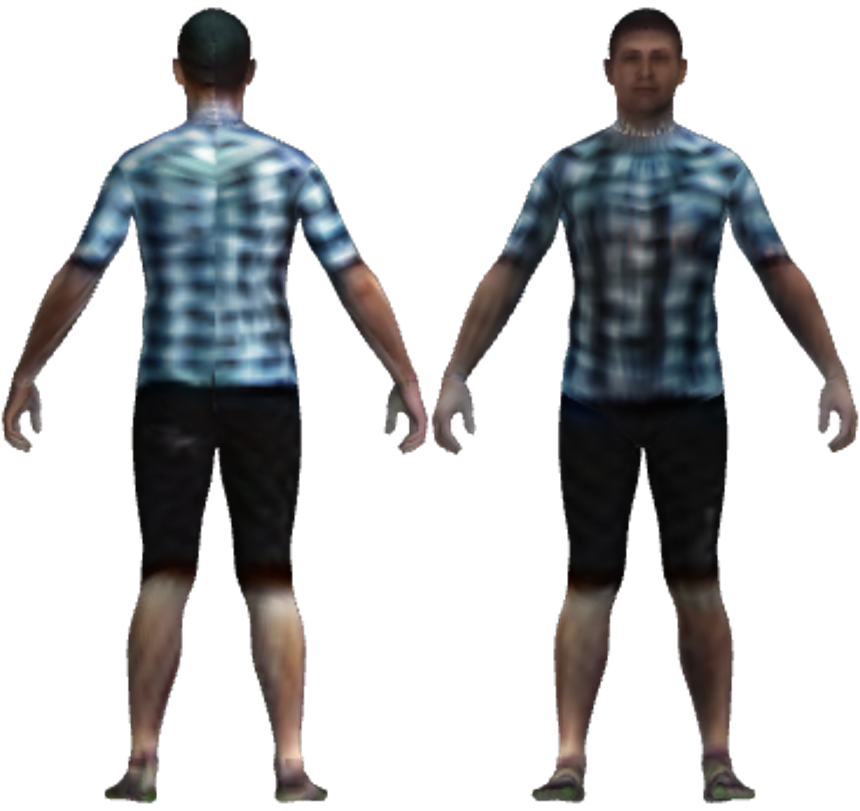}\\

     
    \interpfigm{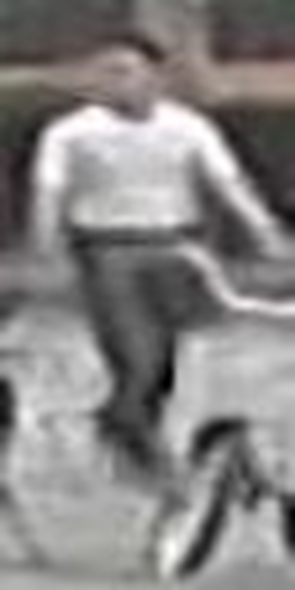}&
    \interpfigt{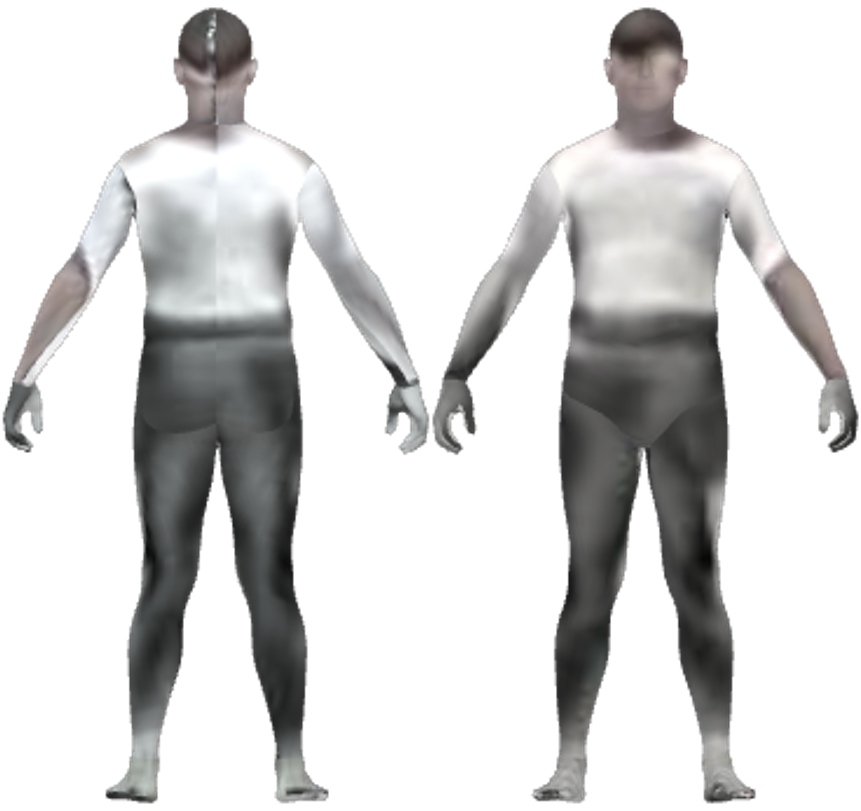}&
     \interpfigt{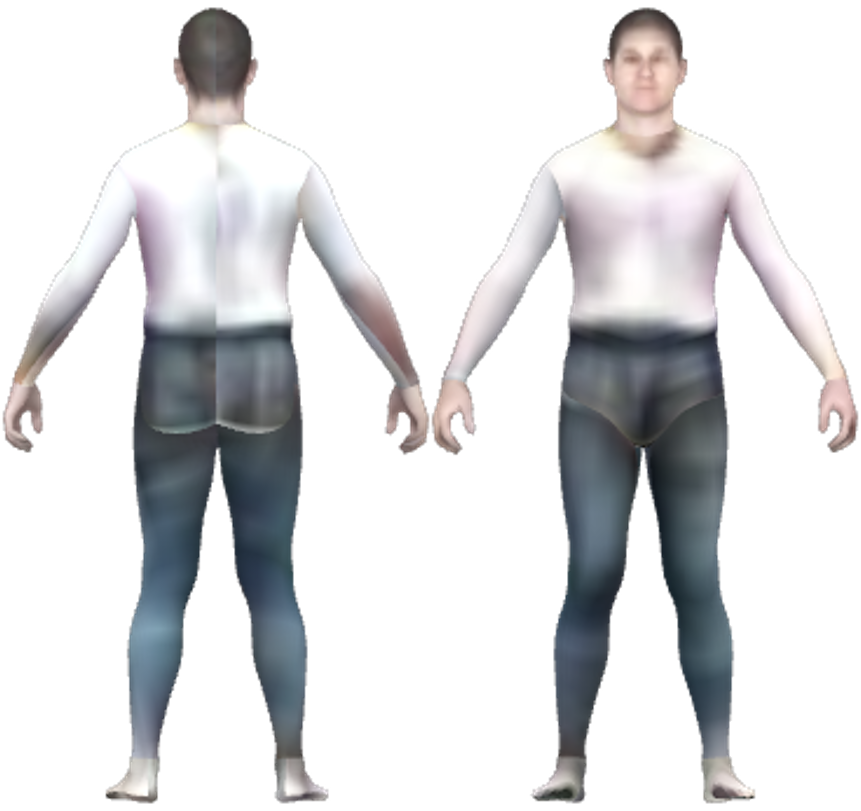}&
     \interpfigt{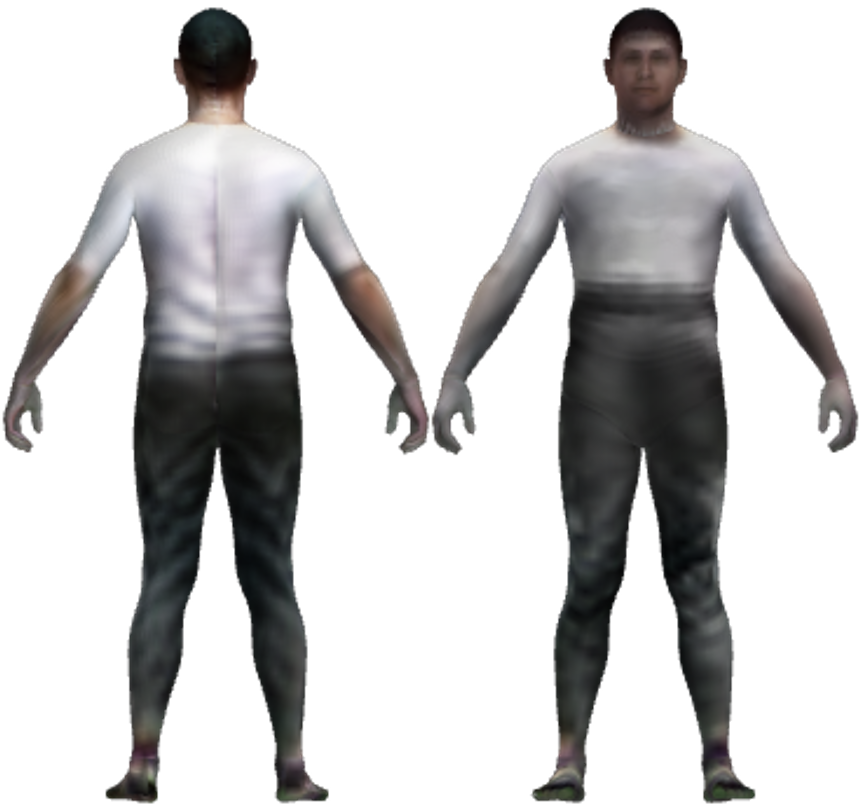}&
     \interpfigt{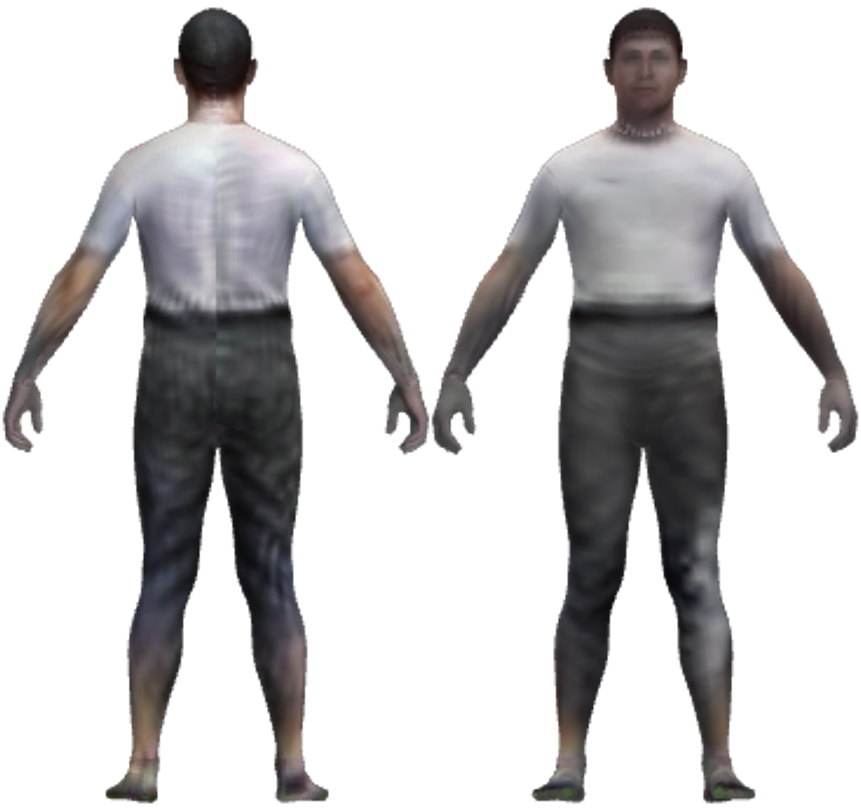}\\
       \interpfigm{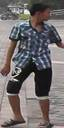}&
    \interpfigt{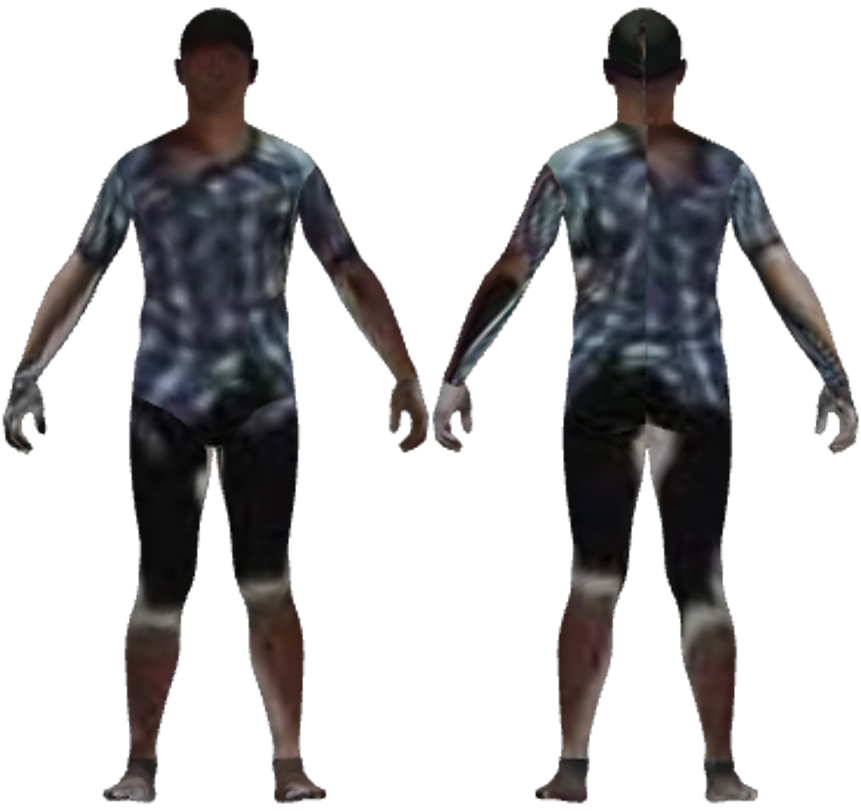}&
     \interpfigt{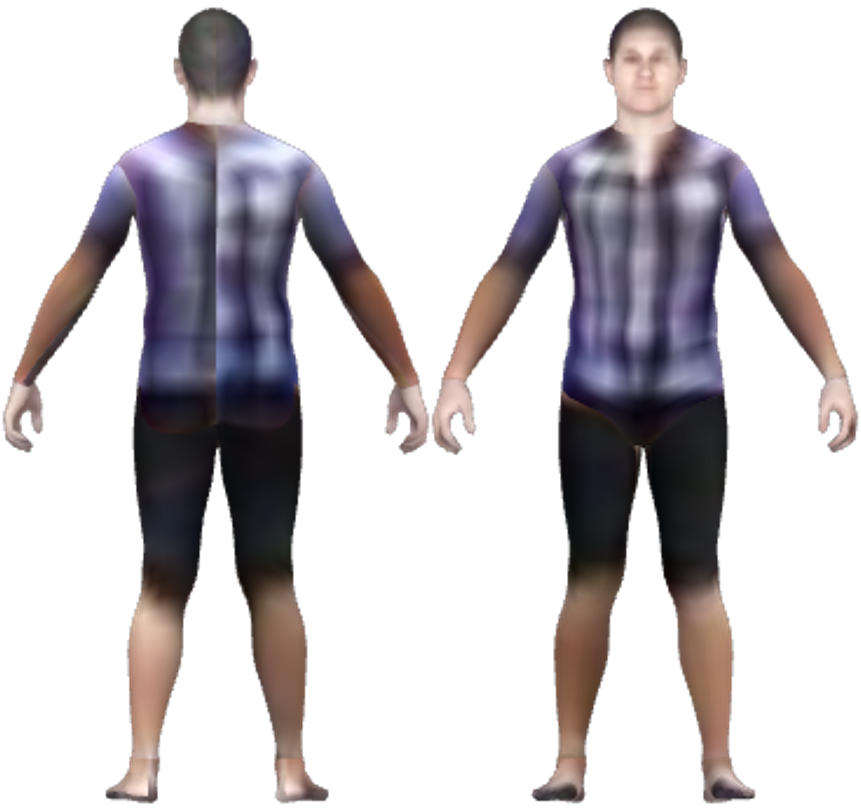}&
     \interpfigt{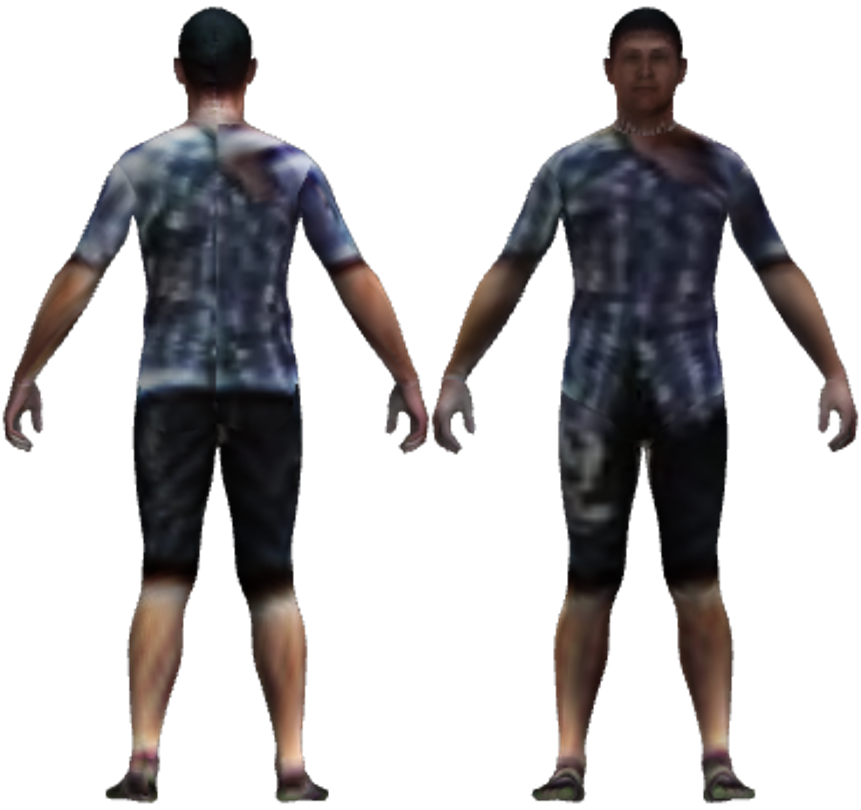}&
     \interpfigt{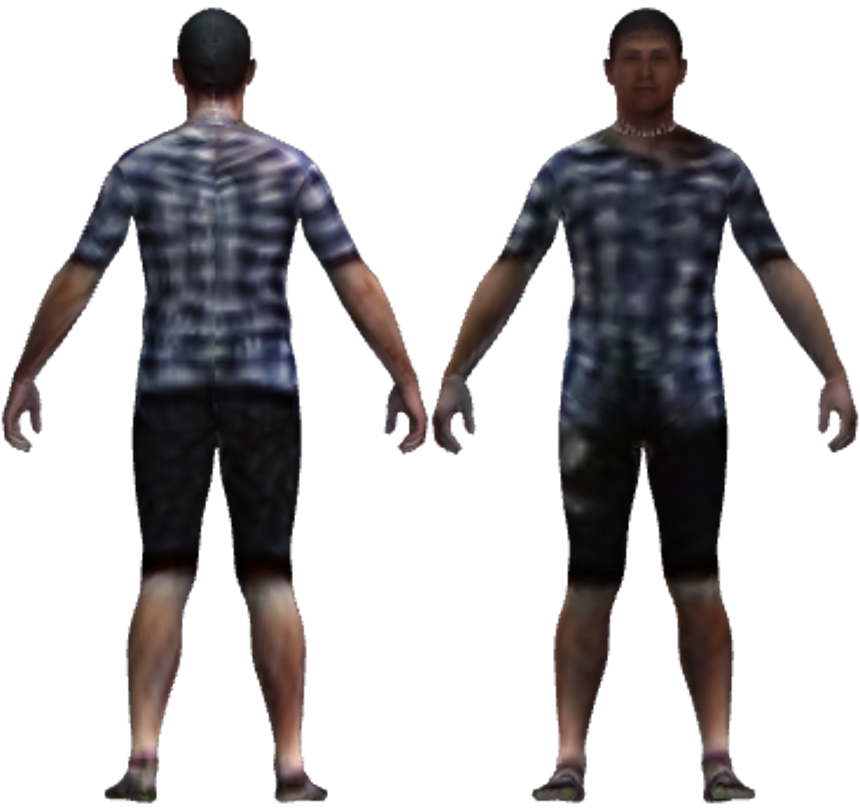}\\

Input & HPBTT \cite{zhao2020human} & RSTG \cite{wang2019re} &  Texformer \cite{xu2021texformer}   & Ours  \\
 
 \end{tabular}
 \caption{Qualitative results of our method and state-of-the-art competing methods on the Market-1501 dataset. }
 \label{fig:final_res}
\end{figure*}

\newcommand{\interpfigi}[1]{\includegraphics[height=2.6cm]{#1}}

\begin{figure*}[h]
  \centering
  \setlength\tabcolsep{1.8pt}
  \renewcommand{\arraystretch}{0.48}
  \begin{tabular}{cccccc}
     \interpfigi{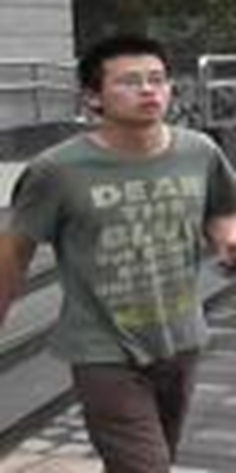}&
     \interpfigi{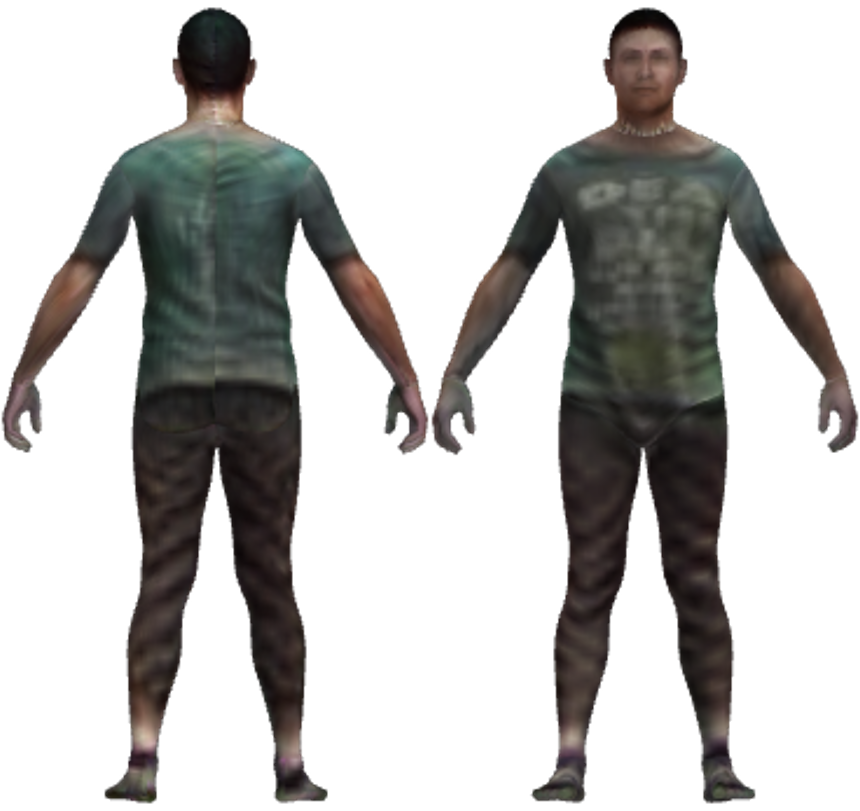}&
    \interpfigi{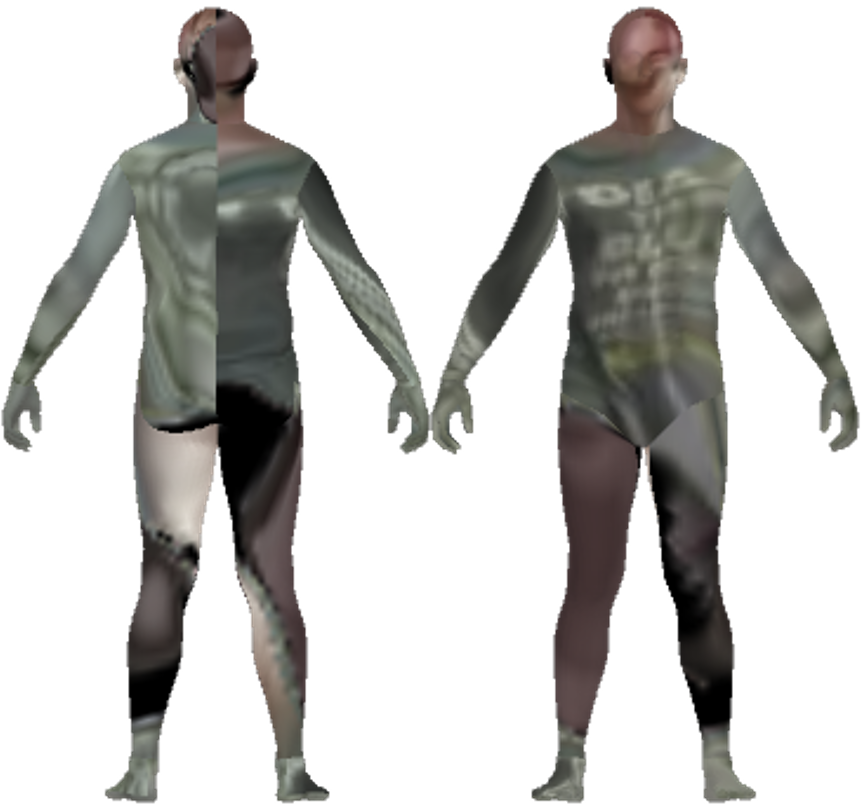}&
    \interpfigi{Figures/final_comparisons/GT2.png}&
     \interpfigi{Figures/final_comparisons/Ours-GT2.png}
     &\interpfigi{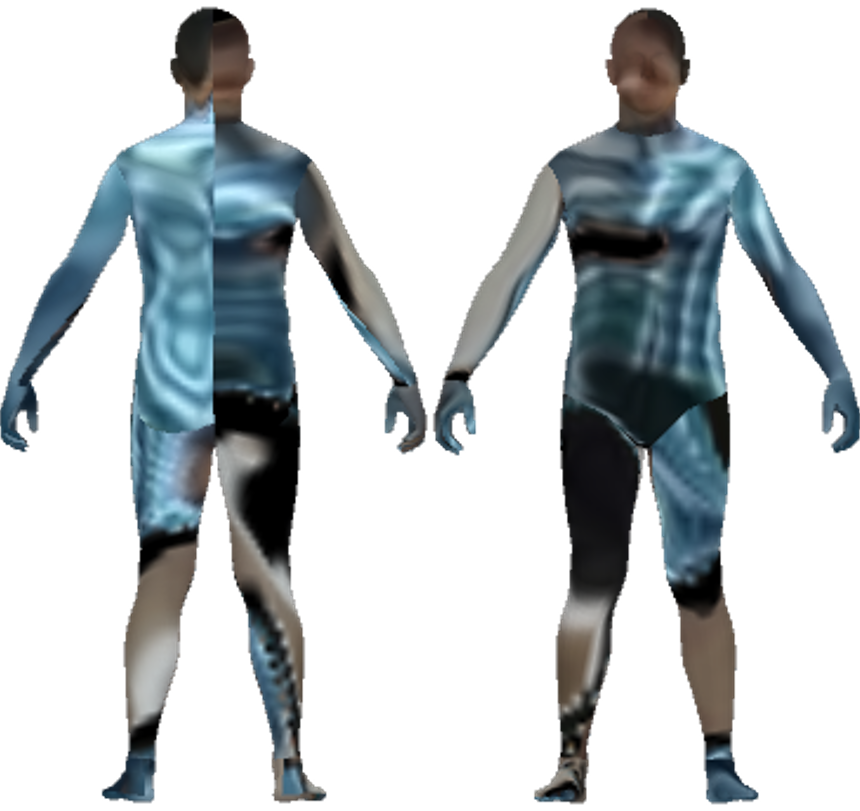}\\ 
    \interpfigi{Figures/final_comparisons/GT3.png}&
    \interpfigi{Figures/final_comparisons/Ours-GT3.png}&
    \interpfigi{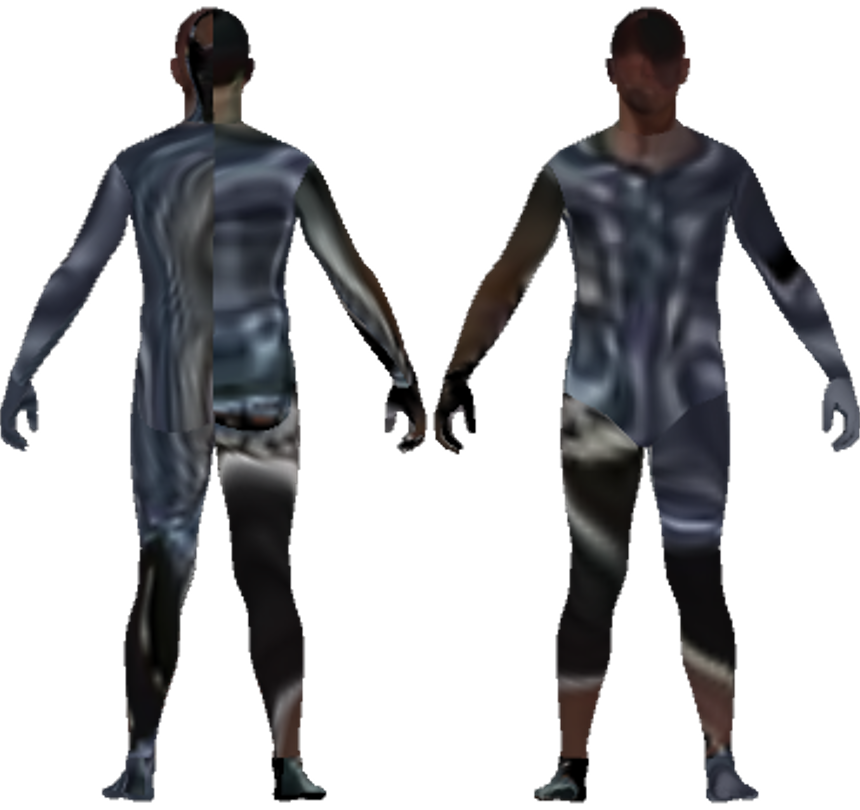}&
    \interpfigi{Figures/final_comparisons/GT5.png}&
    \interpfigi{Figures/final_comparisons/Ours-GT5.png}&
    \interpfigi{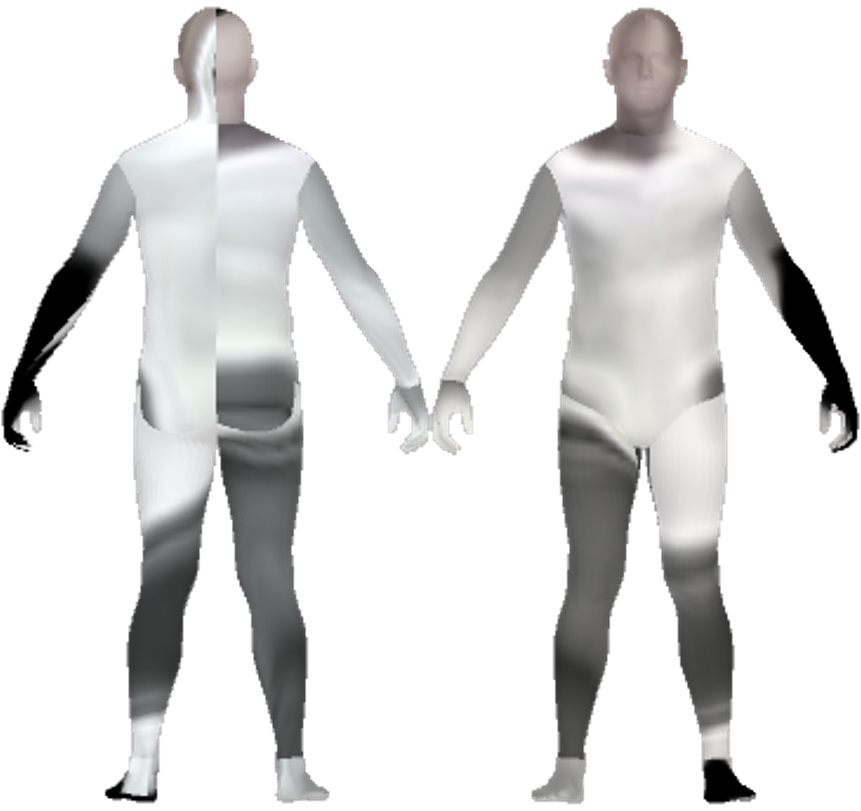} \\
     
     
Input & Ours   & CoordInpaint \cite{grigorev2019coordinate} & Input & Ours   & CoordInpaint \cite{grigorev2019coordinate}   \\
 
 \end{tabular}
 \caption{Qualitative results of our method and Coordinate-based texture inpainting \cite{grigorev2019coordinate} on the Market-1501 dataset. }
 \label{fig:sup_comp_coord}
\end{figure*}

\begin{figure*}[h]
  \centering
  \setlength\tabcolsep{1.8pt}
  \renewcommand{\arraystretch}{0.48}
  \begin{tabular}{ccccc}
  \interpfigm{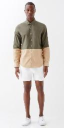}&
    \interpfigt{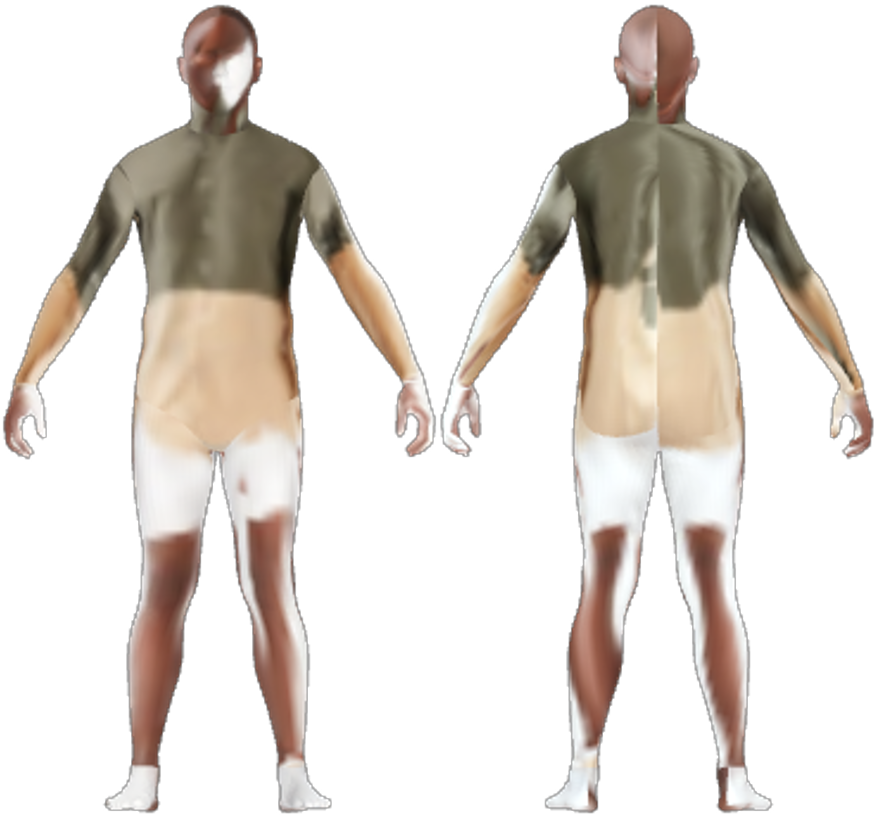}&
    \interpfigt{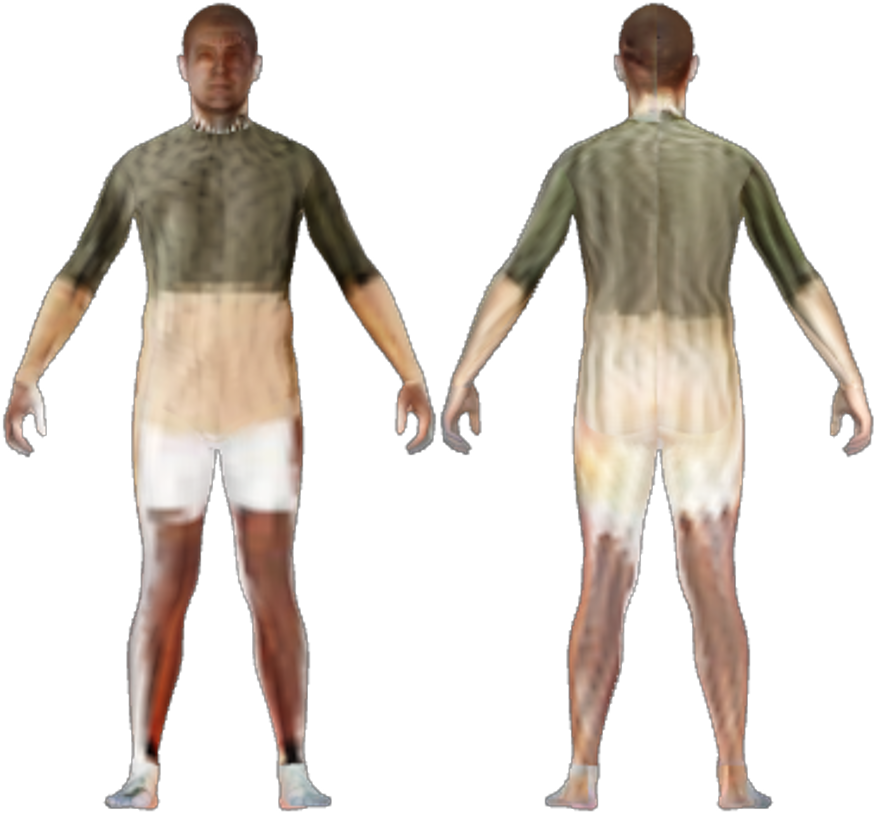}&
    \interpfigt{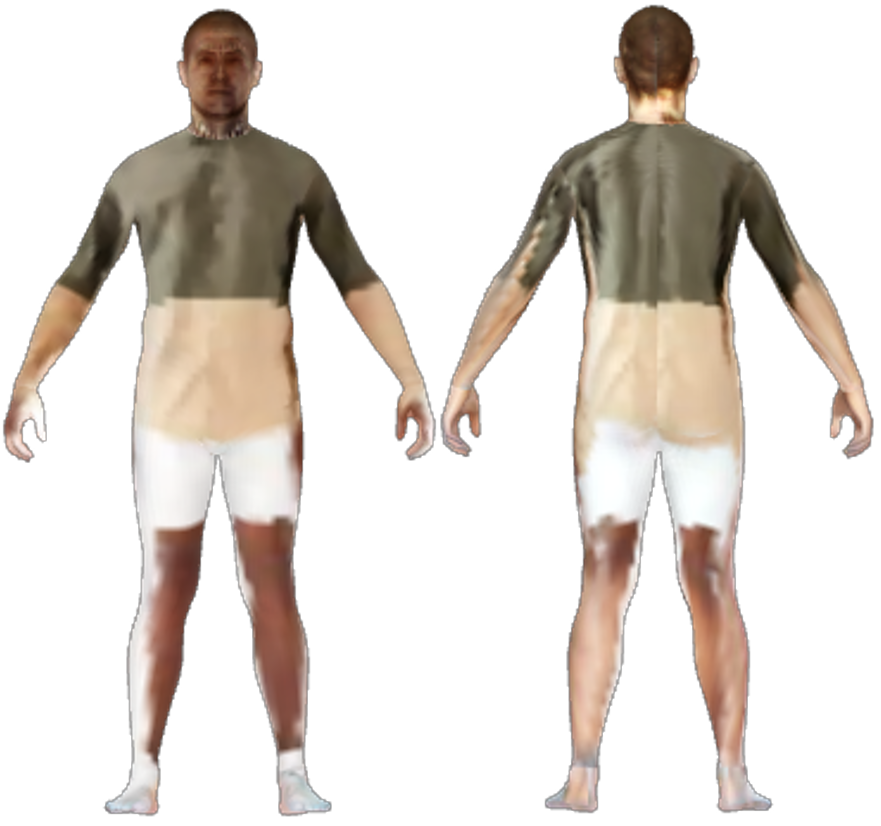}\\

  \interpfigm{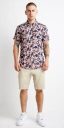}&
    \interpfigt{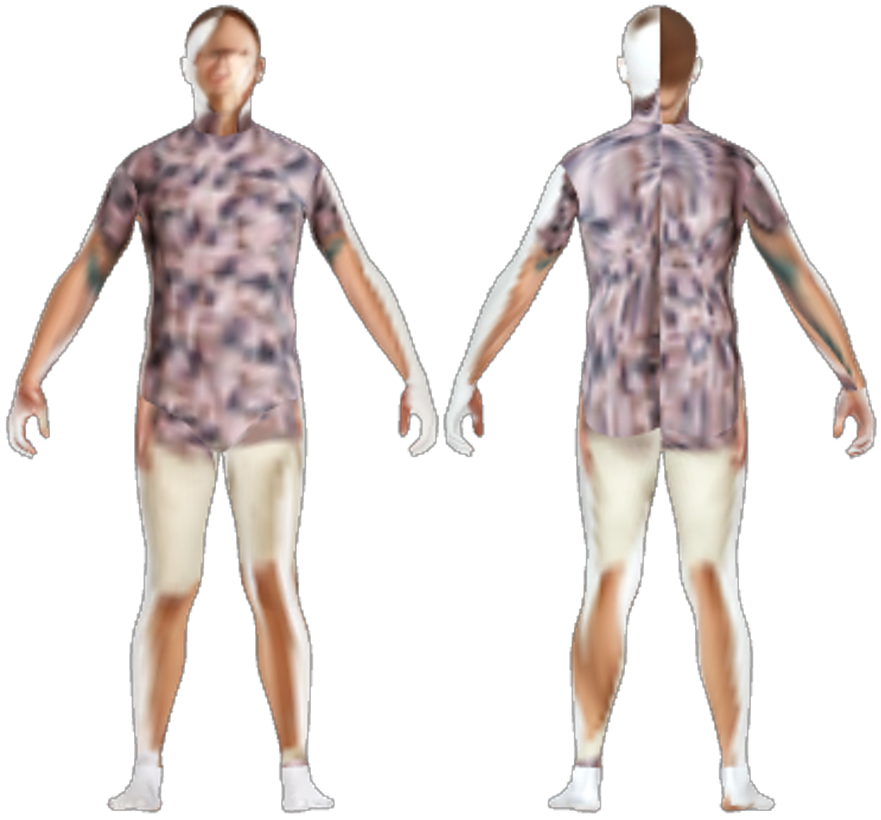}&
    \interpfigt{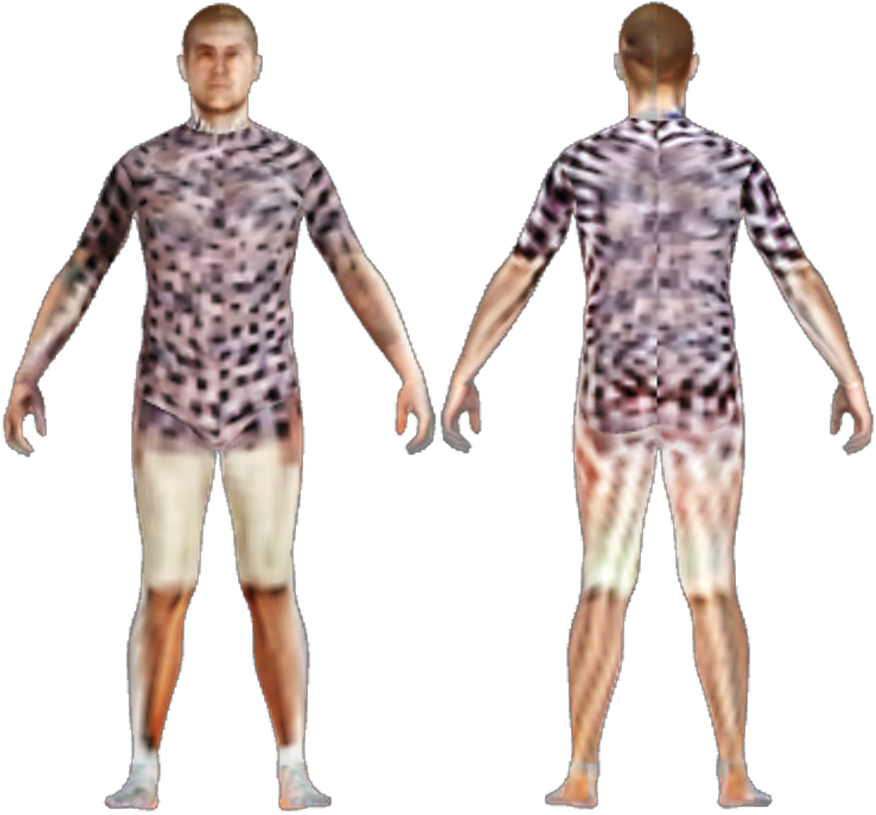}&
    \interpfigt{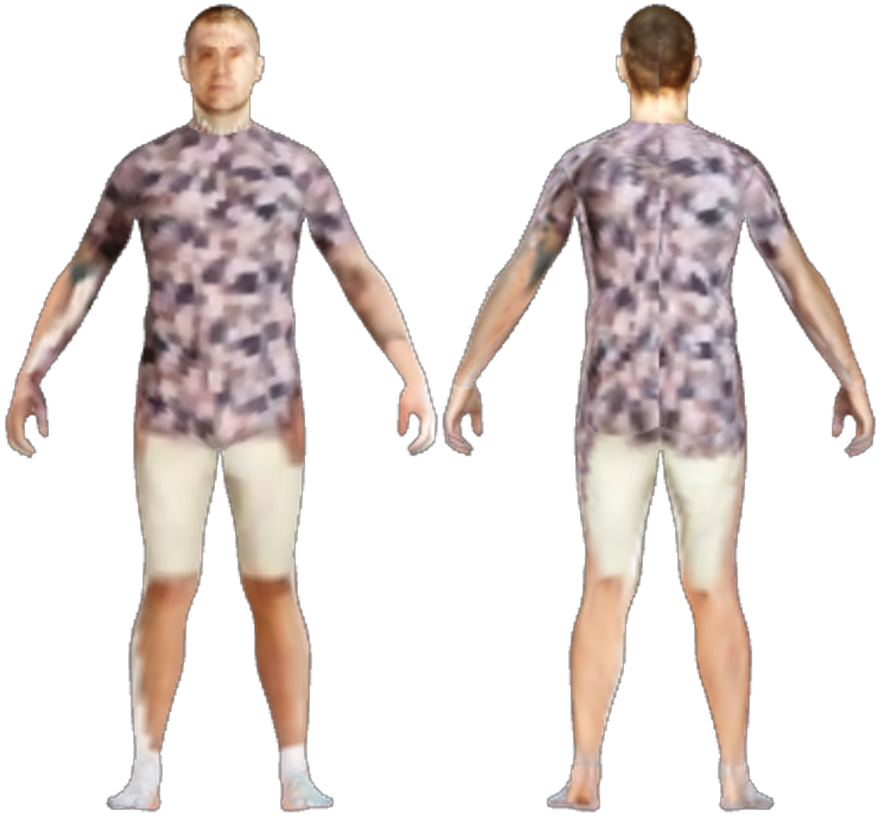}\\ 

  \interpfigm{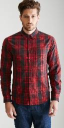}&
    \interpfigt{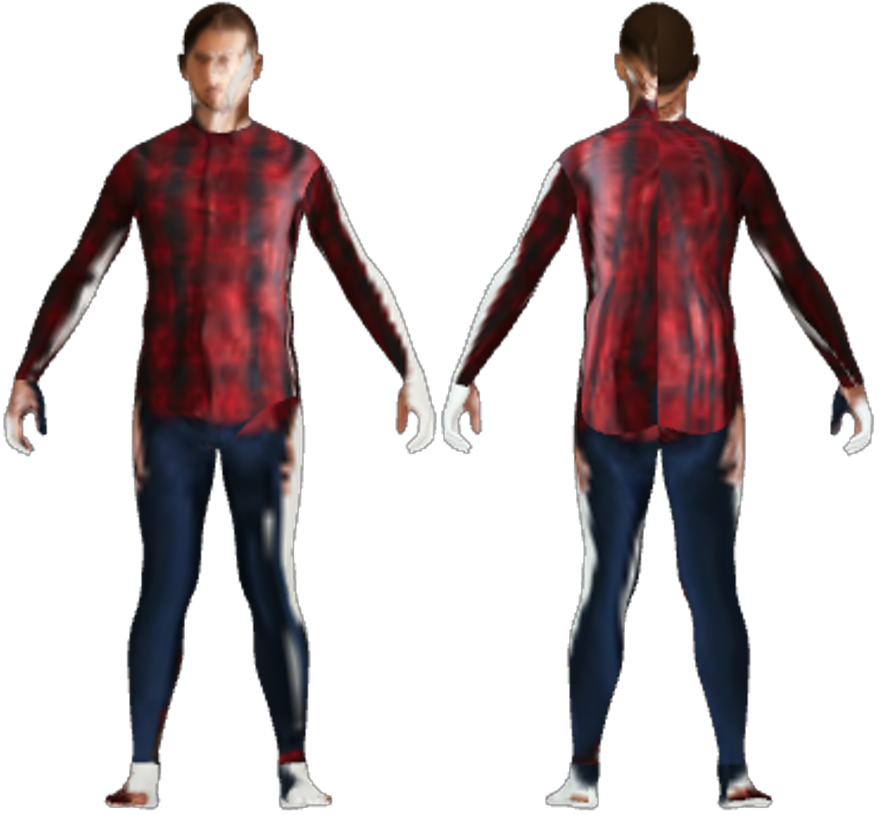}&
    \interpfigt{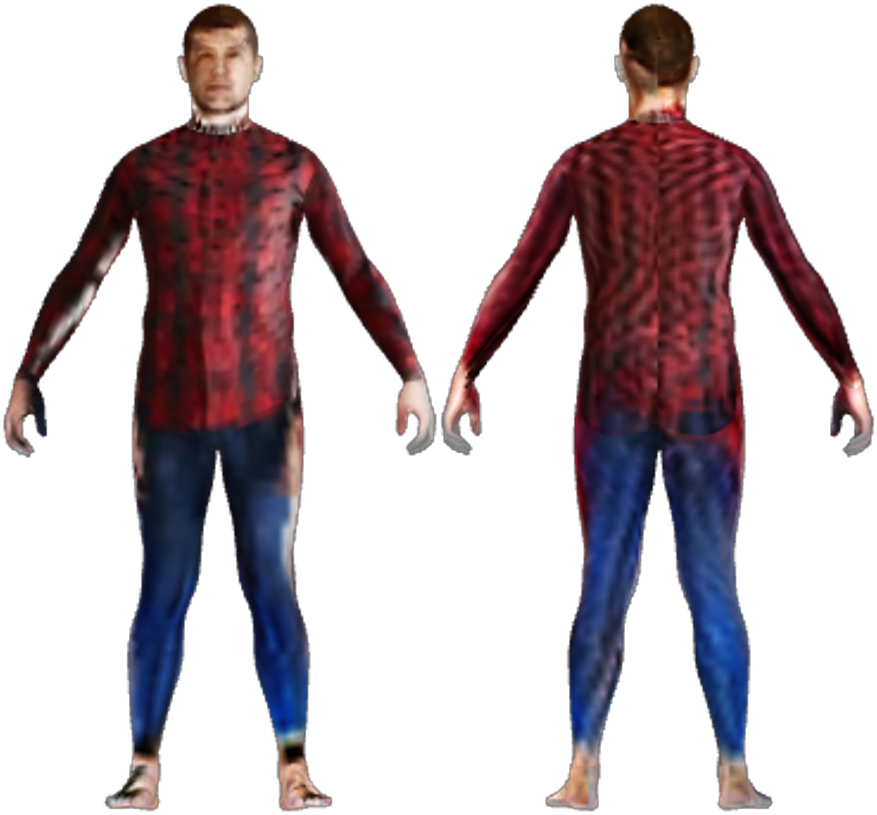}&
    \interpfigt{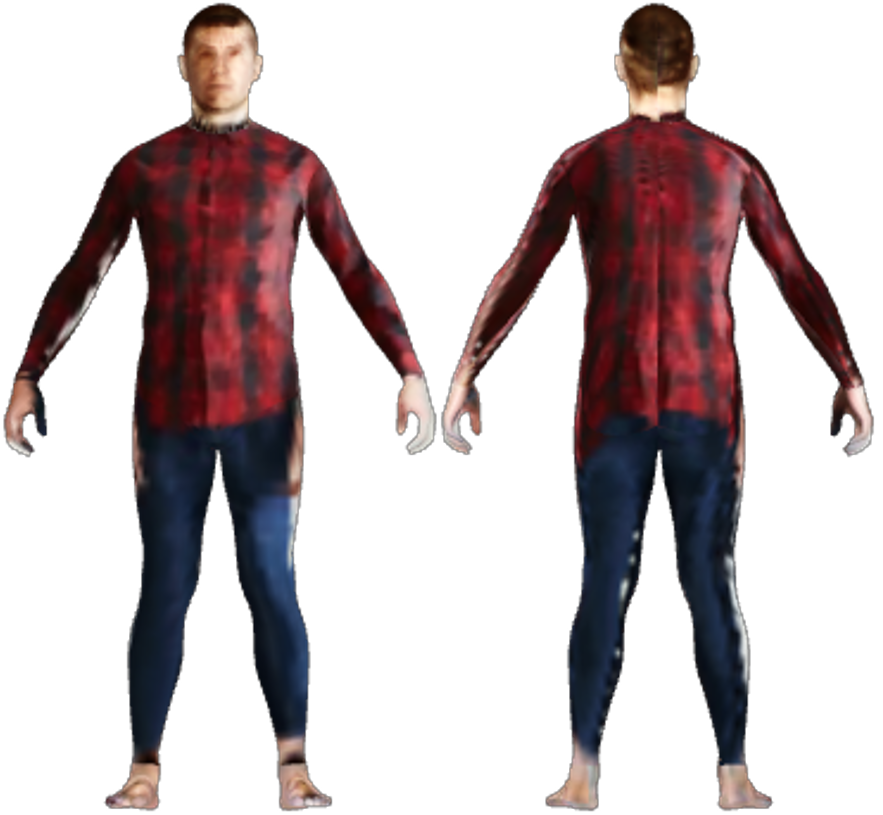}\\

Input & HPBTT \cite{zhao2020human} & Texformer \cite{xu2021texformer}   & Ours  \\
 
 \end{tabular}
 \caption{Qualitative results of our method and state-of-the-art competing methods on the DeepFashion dataset. }
 \label{fig:final_res_fashion}
\end{figure*}


We compare our method against the state-of-the-art 3D human texture estimation methods: HPBTT \cite{zhao2020human}, RSTG \cite{wang2019re},  TexGlo \cite{xu20213d}, and Texformer \cite{xu2021texformer}.
Quantitative results for Same-view (SV) and Novel-view (NV) are presented in Table~\ref{tab:final}. We calculate the results of NV of the methods with released models of RSTG and Texformer. 
First, we compare the methods on the Market-1501 dataset.
Since HPBTT uses a different data split, we train it with our split with their released code.
HPBTT~\cite{zhao2020human} outputs texture flow with a regular convolutional neural network that takes body segmentation and pose as input. Since the method outputs texture flow, the colors match the input image, which results in a relatively good performance on SSIM and LPIPS metrics for the same view. On LPIPS, HPBTT even achieves the second-best result. However, its results are pretty poor when measured with CosSim and CosSim-R metrics due to the artifacts in the final renderings. RSTG~\cite{wang2019re} and  TexGlo~\cite{xu20213d} output results in fewer artifacts, which leads to relatively better performance on CosSim and CosSim-R metrics. On the other hand, their results do not match the input images in colors and fine details, which causes significantly worse SSIM and LPIPS scores. Texformer~\cite{xu2021texformer} is closest to our work with good metrics overall. In our work, we significantly improve their results. Our proposed method achieves consistently better results than the competing methods on all metrics for both same and novel view evaluations with slightly more parameters than Texformer and significantly fewer parameters than the others. 

Qualitative results of our method and competing methods are presented in Fig. \ref{fig:final_res}. HPBTT model outputs results with many artifacts, especially in the invisible regions.
RSTG does not suffer from these severe artifacts; however, the method outputs blurry predictions lacking fine details. Texformer achieves better results than HPBTT and RSTG but still does not achieve high fidelity to the input images, especially on the challenging patterns, e.g., check shirts. Finally, our method achieves high-quality results with fine details, high-fidelity colors, and texture patterns.


We also compare with coordinate-based texture inpainting for pose-guided human image generation \cite{grigorev2019coordinate}. CoordInpaint has two pipelines. First, dense poses are estimated by the Dense Pose method \cite{guler2018densepose}, and then they are converted to SMPL coordinates using a predefined mapping (provided with the
DensePose). They are used in both pipelines. In the first pipeline, a complete body texture is obtained through an inpainting network. 
The output of the first pipeline is a texture map that can be used to render humans with different poses.
In the second pipeline, images are rendered for a target pose and further processed in the image space. 
The second pipeline, therefore, does not output a 3D model but refines the results in image space.
Hence, our method is comparable with the output of the first pipeline. We provide the comparison results in Fig.~\ref{fig:sup_comp_coord}. 
Coordinpaint does not achieve good results after its first pipeline. Since their model is proposed for two pipelines, we do not include their work in our main comparisons.

Next, we compare methods on the DeepFashion dataset in Table \ref{tab:final}. We train HPBTT~\cite{zhao2020human}, Texformer ~\cite{xu2021texformer}, and our method on this dataset since they are the top-3 methods from the Market-1501 dataset.
As shown in Table~\ref{tab:final}, our method achieves an even more significant improvement on this dataset.
HPBTT slightly achieves a better score on the NV SSIM score. 
HPBTT's SSIM scores are also high on the Market-1501 dataset, and that is because HPBTT outputs texture flow and pixel colors that match the input.
However, it is also shown that the SSIM score sometimes prefers blurred images that match better at the pixel level, and LPIPS is shown to be a better metric that matches human perception~\cite{zhang2018unreasonable}.
However, all the other metrics are poor due to the artifacts in the final renderings.
As shown in Fig.~\ref{fig:final_res_fashion}, the same behavior is observed there as HPBTT outputs blurred images for novel views.
Our method achieves significantly better LPIPS scores.
Our method also shows significant improvements over Texformer.
It may be because the dataset has more variations in the poses, and our contributions achieve significant improvements on this challenging dataset. 

\begin{table*}[t]
\caption{Ablation Study on Market-1501 dataset.}
\label{tab:abl}
\centering
\renewcommand\tabcolsep{12pt}
\begin{tabular}{|l|cc|cc|cc|cc|}
\hline
& \multicolumn{2}{c}{SSIM \ \ \ $\Uparrow$} & \multicolumn{2}{|c}{LPIPS \ \ \ $\Downarrow$} & \multicolumn{2}{|c}{CosSim \ \ \ $\Uparrow$}  & \multicolumn{2}{|c|}{CosSim-R \ \ \ $\Uparrow$} \\
\hline
& SV & NV & SV & NV & SV & NV & SV & NV \\
\hline
BL (Baseline) &  0.7422 & 0.6535 & 0.1154  & 0.2040 & 0.5747 & 0.4943 & 0.5422 & 0.4736\\
\hline
BL + Conv Refine & 0.7196 & 0.6520 & 0.1293 & 0.2052 & 0.5701 & 0.4976 & 0.5343 & 0.4758 \\
BL + Deformable Refine  & 0.7490 & 0.6511 & \textbf{0.1038} & 0.2046 & \textbf{0.5887} & \textbf{0.4976} & \textbf{0.5568} & \textbf{0.4775} \\
BL + URL &  \textbf{0.7560} & \textbf{0.6540} & 0.1053 & \textbf{0.2020} & 0.5813 & 0.4802 & 0.5481 & 0.4767 \\
BL + Cycle & 0.7448 & 0.6503 & 0.1107 & 0.2027 & 0.5837 & 0.4968 & 0.5502 & 0.4761 \\
\hline
\end{tabular}
\end{table*}

\begin{figure}[]
  \centering
  \setlength\tabcolsep{3.08pt}
  \renewcommand{\arraystretch}{0.48}
  \begin{tabular}{ccc}
       \interpfigi{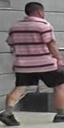}&
    \interpfigi{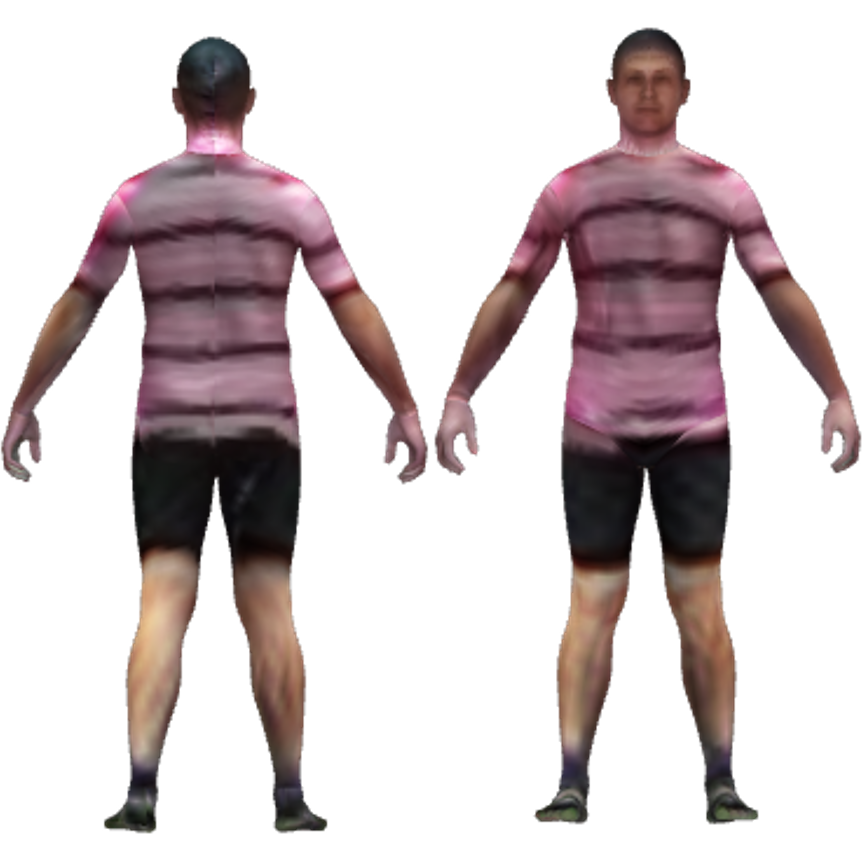}&
     \interpfigi{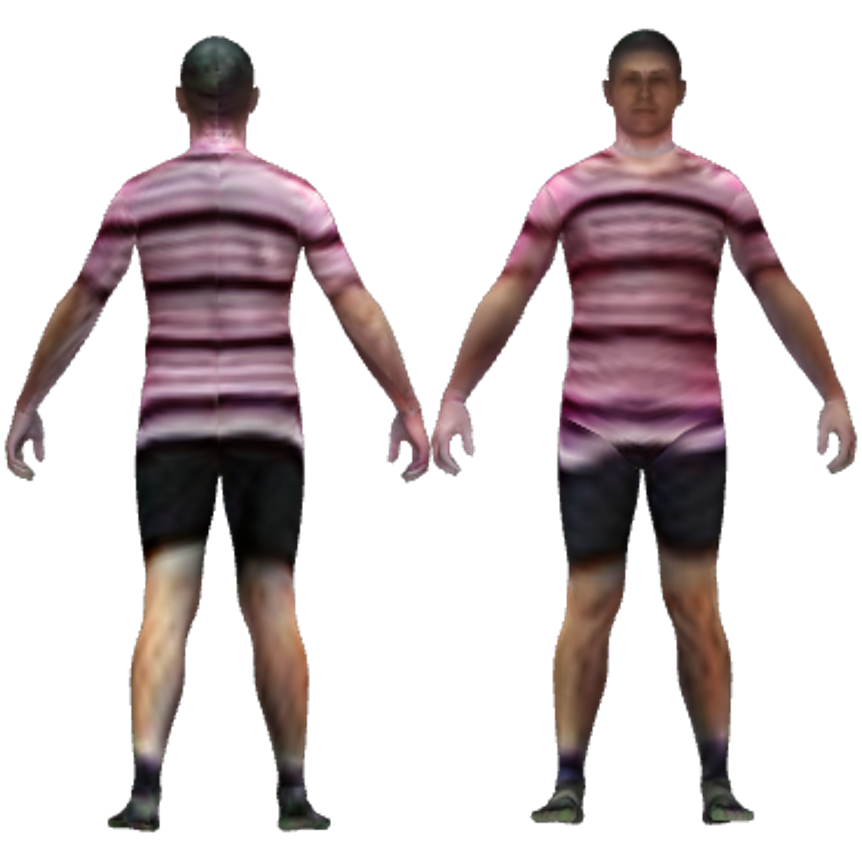}\\
       Input & BL & BL+Refine\\
     \interpfigi{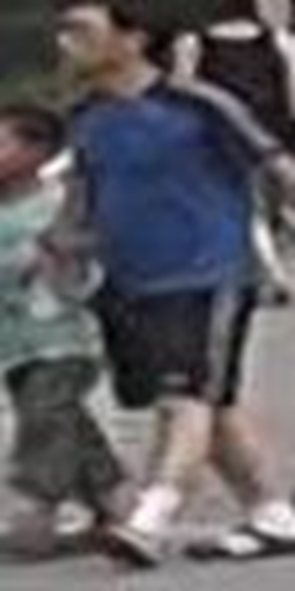} & 
     \interpfigi{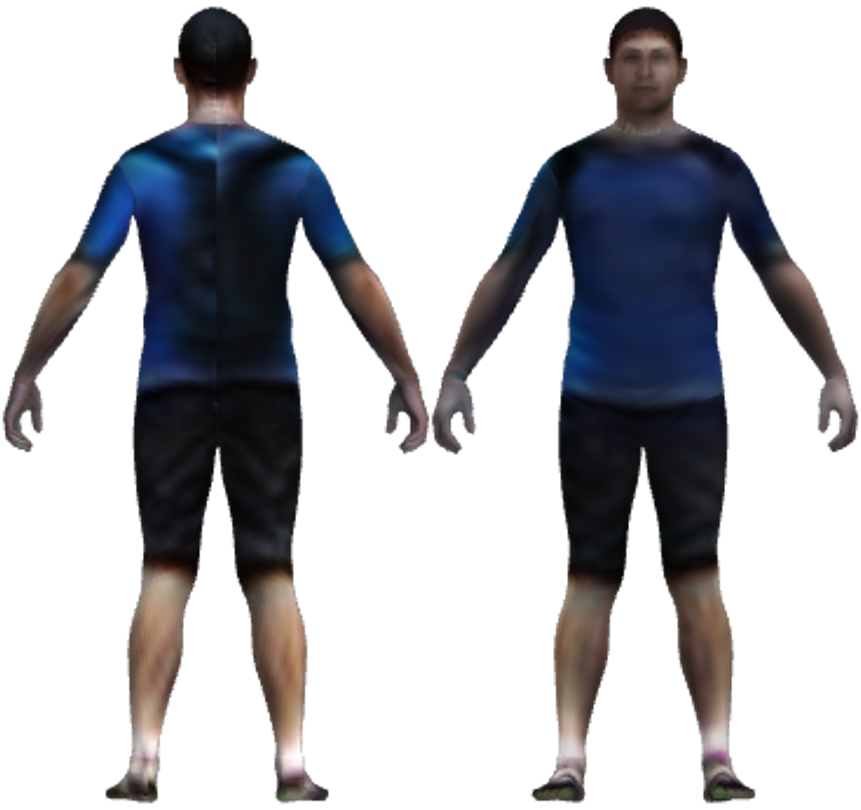}&
     \interpfigi{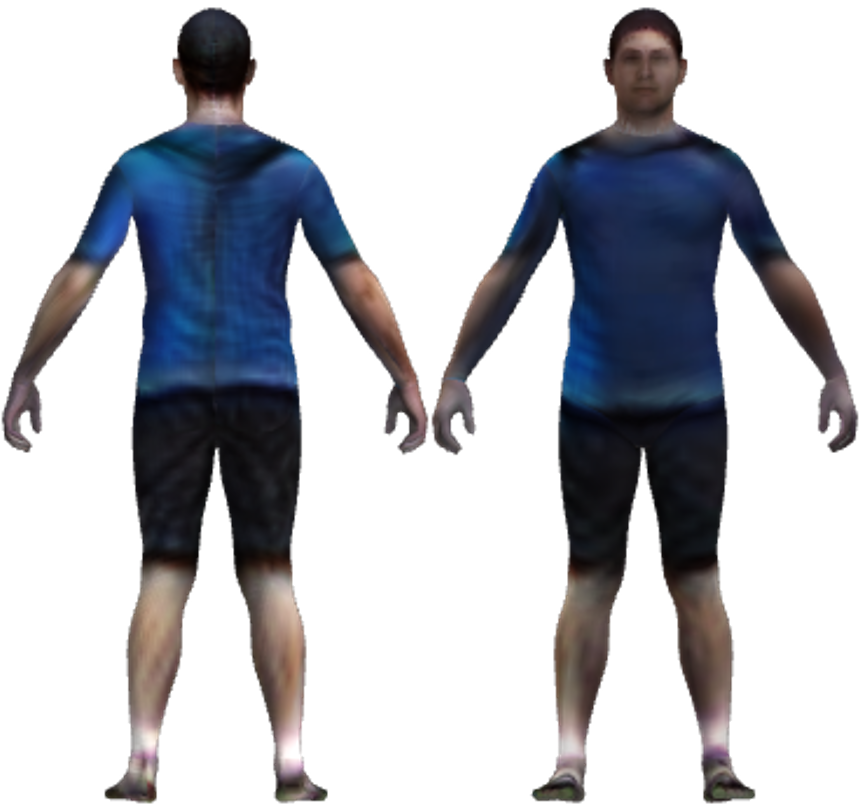}\\
      Input  & BL & BL+URL  \\
     \interpfigi{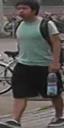} & 
     \interpfigi{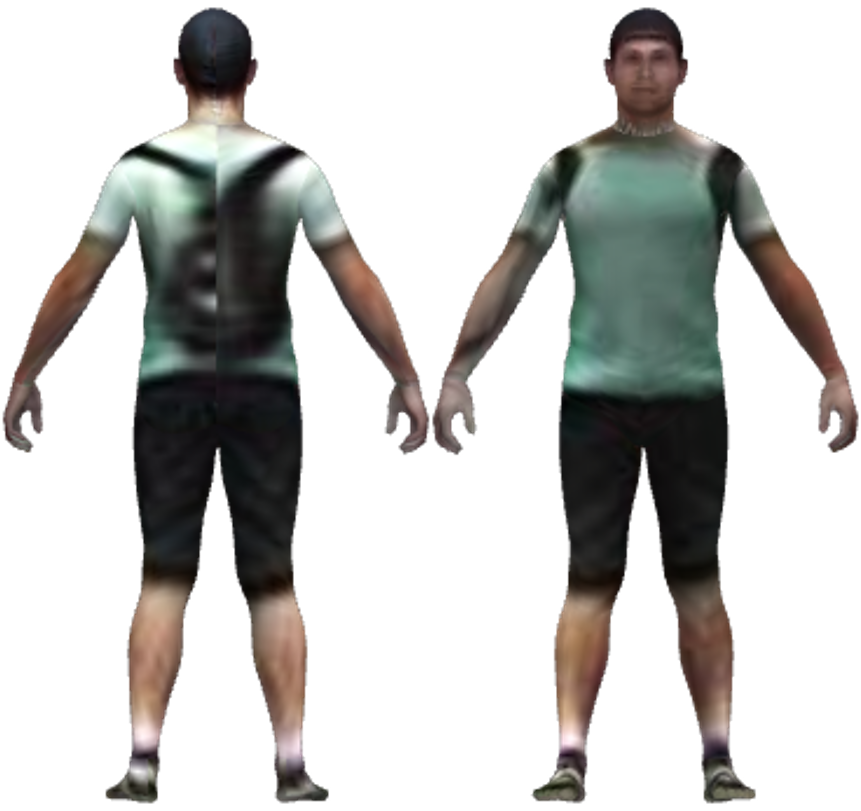}&
     \interpfigi{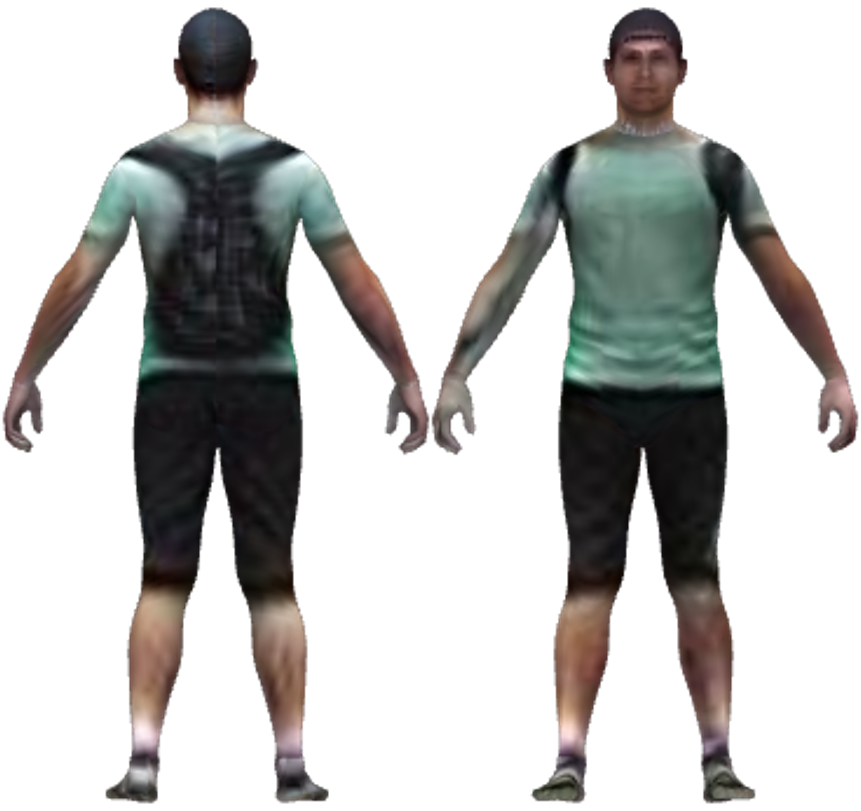}\\
     Input  & BL & BL+Cycle  \\
     \end{tabular}
     \caption{ Qualitative results of Ablation Study.}
     \label{fig:abl}
\end{figure}

\subsection{Ablation Study}

We conduct an ablation study to show the improvements of each proposed contribution as provided in Table~\ref{tab:abl} and Fig.~\ref{fig:abl}. Our baseline model (BL) only includes the deep network from Fig.~\ref{fig:overall_frame} without the final branch, which is the deformable convolution-based module. BL is only trained with base losses described in Sec.~\ref{sec:baseloss}. We then test each proposed change separately. First, we add a deformable-based refinement module to the architecture (BL+Refine). Next, we experiment with proposed loss objectives by adding uncertainty-based reconstruction loss (BL+URL) and cycle consistency loss (BL+Cycle).

 

\begin{figure}
    \centering
    \includegraphics[width=1\linewidth]{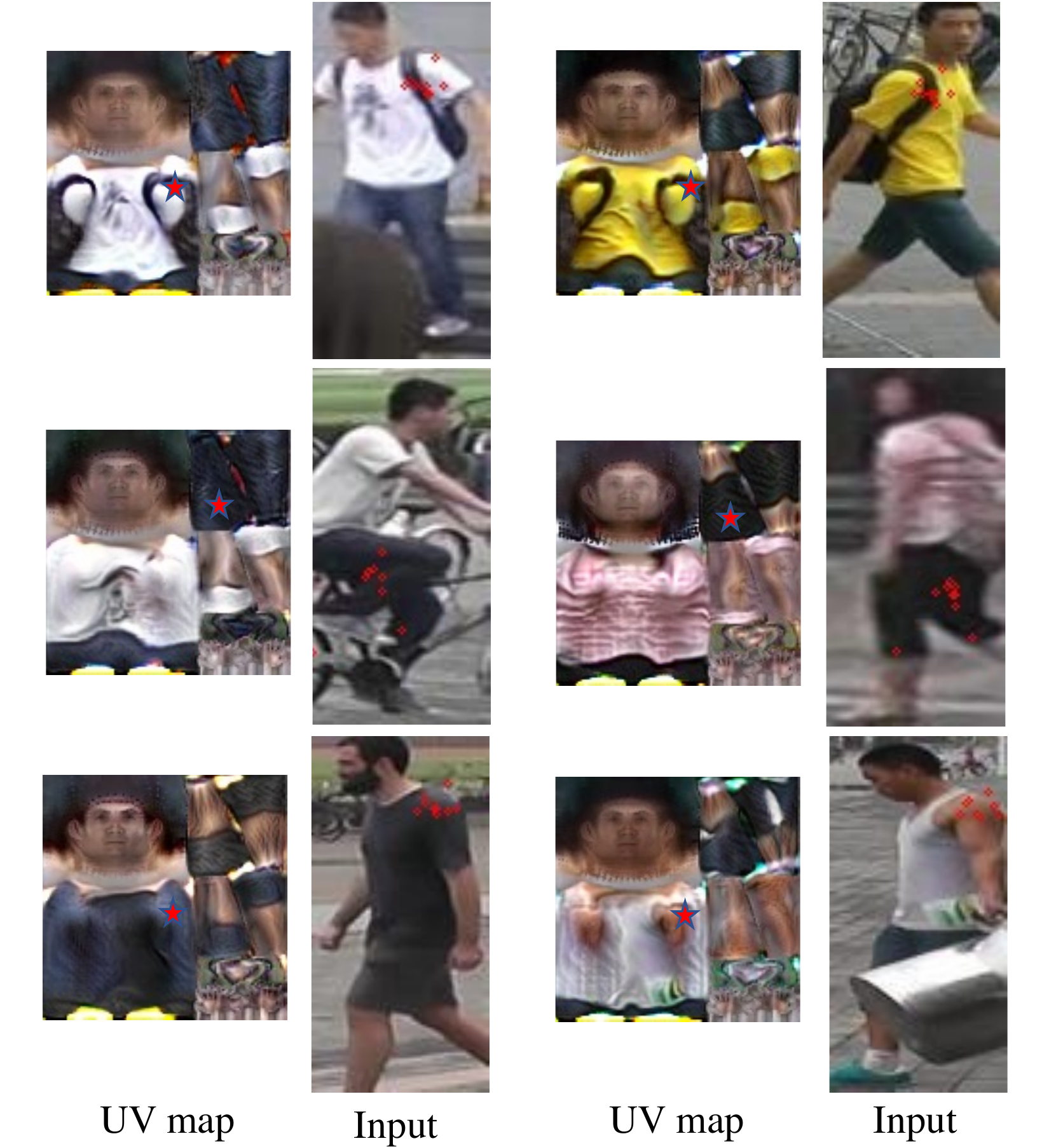}
    \caption{Visualization of offsets in input images.}
\label{fig:offsets}
\end{figure}

Adding a deformable convolution-based refinement module improves all metrics. 
To measure the effectiveness of the proposed refinement module, we experiment if the improvements come from the increased capacity (additional learning parameters). To test that, we experiment with a shallow UNET architecture as the refinement module, which is convolution-based. As shown in Table \ref{tab:abl}, the additional layers do not improve the results.
As shown in Fig.~\ref{fig:abl}, the refinement module improves challenging scenarios where the shirt stripes have superior quality.
In Fig. \ref{fig:offsets}, we visualize the corresponding offsets of deformable convolution for marked points from the UV map.
We mark the same coordinates for the two examples in each row. 
As shown in the input image, the offsets are in different locations in the input based on the content.
The texture estimation task fits well with deformable convolution.
URL significantly improves the SSIM and LPIPS metrics by providing pixel-level reconstruction loss. The second example in Fig.~\ref{fig:abl} has superior color fidelity to the input image.
The third row of Fig.~\ref{fig:abl} shows that the cycle consistency loss significantly improves the prediction of invisible regions.
As can be seen, with the cycle consistency loss, a more realistic backpack is predicted by only seeing its straps.

\section{Conclusion}
\label{sec:conc}


We propose a framework to refine 3D human textures estimation from a single image. We use a deformable convolution-based refinement module to sample an input image for better quality adaptively. We also introduce an uncertainty-based reconstruction and novel cycle consistency losses responsible for our high-fidelity texture estimation. We show several qualitative and quantitative improvements compared to the state-of-the-art methods.
We hope our work will inspire future research to test their texture inferences for view generalization by evaluating novel inferences.



\section*{Acknowledgement} 
A. Dundar acknowledges the support of the Marie Skłodowska-Curie Individual Fellowship.

{
\bibliographystyle{ieee}
\bibliography{egbib}
}

\end{document}